\RequirePackage{fix-cm}
\documentclass[smallextended]{svjour3}       
\smartqed  
\usepackage[utf8]{inputenc}
\usepackage{graphicx}
\usepackage[english]{babel}
\usepackage{amsmath}
\usepackage{amssymb}
\usepackage{hyphenat} 
\usepackage{algorithmic}
\usepackage{algorithm}
\usepackage{subfig}
\usepackage{microtype}
\begin{document}

\title{Incorporating Expert Prior Knowledge into Experimental Design via Posterior Sampling}


\author{Cheng Li         \and
        Sunil Gupta      \and
        Santu Rana       \and
        Vu Nguyen        \and
        Antonio Robles-kelly \and
        Svetha Venkatesh
}


\institute{Cheng Li \at
              National University of Singapore, Singapore; Applied Artificial Intelligence Institute (A2I2), Deakin University, Australia \\
              \email{licheng@comp.nus.edu.sg}     
           \and
           Sunil Gupta \at
              Applied Artificial Intelligence Institute (A2I2), Deakin University, Australia \\
              \email{sunil.gupta@deakin.edu.au} 
           \and
           Santu Rana \at
              Applied Artificial Intelligence Institute (A2I2), Deakin University, Australia \\
              \email{santu.rana@deakin.edu.au} 
           \and
           Vu Nguyen \at
              University of Oxford, United Kingdom \\
              \email{vu@robots.ox.ac.uk} 
           \and
           Antonio  Robles-kelly\at
              Deakin University, Australia \\
              \email{antonio.robles-kelly@deakin.edu.au} 
           \and
           Svetha Venkatesh\at
              Applied Artificial Intelligence Institute (A2I2), Deakin University, Australia \\
              \email{svetha.venkatesh@deakin.edu.au} 
}

\date{Received: date / Accepted: date}

\maketitle

\begin{abstract}
Scientific experiments are usually expensive due to complex \texttt{experi\-mental} preparation and processing. Experimental design is therefore involved with the task of finding the optimal experimental input that results in the desirable output by using as few experiments as possible. Experimenters can often acquire the knowledge about the location of the global optimum. However, they do not know how to exploit this knowledge to accelerate experimental design. In this paper, we adopt the technique of Bayesian optimization for experimental design since Bayesian optimization has established itself as an efficient tool for optimizing expensive black-box functions. Again, it is unknown how to incorporate the expert prior knowledge about the global optimum into Bayesian optimization process. To address it, we represent the expert knowledge about the global optimum via placing a prior distribution on it and we then derive its posterior distribution. An efficient Bayesian optimization approach has been proposed via posterior sampling on the posterior distribution of the global optimum. We theoretically analyze the convergence of the proposed algorithm and discuss the robustness of incorporating expert prior. We evaluate the efficiency of our algorithm by optimizing synthetic functions and tuning hyperparameters of classifiers along with a real-world experiment on the synthesis of short polymer fiber. The results clearly demonstrate the advantages of our proposed method.
\end{abstract}

\global\long\def\gv{\given}%
\global\long\def\given{\mid}%

\global\long\def\argmax#1{\underset{_{#1}}{\text{argmax}} }%

\global\long\def\argmin#1{\underset{_{#1}}{\text{argmin}\ } }%

\section{Introduction}
Experimenting advances scientific progress. However, experimenters
usually are annoyed by the allocation of time and resource before
conducting experiments, or how to achieve experimental purposes rapidly
and solidly. \emph{Experimental design} is therefore involved with
finding optimal experimental parameters, or experimental configurations
that result in the best performance by using fewer experiments. In
many cases, experimental design is in essence a black-box optimization
problem as the form of the underlying function is unknown. With the
underlying function $f$, experimental design can be formulated as
a maximization (or minimization) problem 
\begin{equation}
\boldsymbol{x}_{*}=\text{argmax}_{\boldsymbol{x}\in\mathcal{X}}f(\boldsymbol{x}),
\end{equation}
where $\boldsymbol{x}_{*}$ is the global maximizer and $\mathcal{X}$
is the search space of the design parameters or variables $\boldsymbol{x}$.
We assume an optimization setting where the evaluation of the function
evaluation is expensive. This assumption holds fairly typically as
performing experiments in most experimental design domains is costly
in money and/or time. Due to this, we would like to optimize the function
using as few evaluations as possible. Bayesian optimization (BO) is
well known to be an efficient method for optimizing expensive black-box
functions~\cite{brochu_tutorial_2010} and shows competitive performance
in broad applications, such as material search~\cite{xue_accelerated_2016},
experimental design~\cite{brochu_tutorial_2010} and hyperparameter
tuning for machine learning models~\cite{snoek2012practical}.

Typical BO consists of two main steps. The \emph{first} step models
the function relating input variables $\boldsymbol{x}$ to output
$y$ based on existing observations using a probabilistic model. Gaussian
process (GP) is a popular choice used in BO since its posterior has
a tractable analytical form \cite{Rasmussen:2005:GPM:1162254}. The
\emph{second} step uses the posterior GP to construct an acquisition
function which qualifies the information about the next evaluation
and then acquires the maximizer of the acquisition function to be
the next evaluation. The only prior knowledge in BO (using a GP) is
that the unknown function $f$ is smooth and can be modeled using
a GP with an appropriate kernel. Some work has incorporated various
forms of expert prior knowledge about $f$ such as monotonicity~\cite{pmlr-v9-riihimaki10a,LI_etal_ICDM2018}
and unimodality~\cite{Andersen_unimodal_2017}. However, in an optimization
problem, the main goal is to reach close-to-optimum solutions. Any
expert prior knowledge about the optimum location would be extremely
valuable. This kind of expert prior knowledge or hypothesis is indeed
common in practice. For example, domain experts through their experience
over time have some hunches about the region in the parameter space
that yields good product quality (section \ref{subsec:Maximizing-the-Desirable fiber}).
An experienced modeler also has some practical experience about potential
model hyperparameters~\cite{Hutter_2018_hyperparameter,Domhan:2015:SUA:2832581.2832731}.
To the best of our knowledge, no work in the context of Bayesian optimization
has incorporated expert prior knowledge about the location of the
global optimum.

In this paper, we therefore aim to facilitate experimental design
by leveraging expert prior knowledge about the location of the global
optimum in Bayesian optimization. We propose to represent this expert
prior knowledge through a vague probability distribution of a random
variable $\boldsymbol{x}^{*}$ i.e. $\pi(\boldsymbol{x}^{*})$, where
$\boldsymbol{x}^{*}$ is the estimation for $\boldsymbol{x}_{*}$.
Currently, there is no provision in BO algorithms to incorporate the
expert prior $\pi(\boldsymbol{x}^{*})$. Some acquisition functions
used in BO such as entropy search (ES)~\cite{Hennig2012_entropy},
predictive entropy search (PES)~\cite{hernandez2014predictive} and
Thompson sampling (TS)~\cite{pmlr-v70-hernandez-lobato17a} employ
the samples from the posterior distribution $p(\boldsymbol{x}^{*}\gv\mathcal{D}_{n})$
conditioning on the observations $\mathcal{D}_{n}=\{\boldsymbol{x}_{i},y_{i}\}_{i=1}^{n}$,
where $y_{i}$ is a noisy function value at $\boldsymbol{x}_{i}$.
In these cases, $p(\boldsymbol{x}^{*}\gv\mathcal{D}_{n})$ is computed
in the standard GP that implicitly includes a uniform expert prior
on the global optimum. However, it is unknown how to obtain the posterior
distribution of $\boldsymbol{x}^{*}$ after introducing a \emph{non-uniform}
expert prior $\pi(\boldsymbol{x}^{*})$ on the global optimum. In
this paper, we derive a simple yet efficient approach to compute the
posterior distribution $p(\boldsymbol{x}^{*}\gv\mathcal{D}_{n},\pi)$
and then adopt posterior sampling to suggest the next evaluation.
The main contributions in this paper are:
\begin{itemize}
\item We propose an efficient approach to incorporate the expert prior knowledge
about the global optimum into Bayesian optimization to facilitate
experimental design;
\item We theoretically analyze the convergence of the proposed algorithm
and discuss the robustness of incorporating expert prior;
\item We evaluate the efficiency of our algorithm by optimizing synthetic
functions and several real-world applications including hyperparameter tuning and material design.
\end{itemize}

\subsection{Related Work}
\label{subsec:related work}
Researchers and practitioners believe that their experience or understanding
on optimization process can leverage Bayeisan optimization (BO). The prior knowledge about the latent function shape such as monotonicity~\cite{LI_etal_ICDM2018} and unimodility~\cite{Andersen_unimodal_2017} has been considered into BO and demonstrated improvement over the standard BO without that kind of prior knowledge. Another line of related work is to obtain the knowledge from related tasks and then transfer them to the target task. One representative work is multi-task
BO~\cite{NIPS2013_5086}, which transfers the knowledge about the source function
to the optimization of the target function via multi-task GP~\cite{NIPS2007_3189}. Similar ideas such as meta-learning~\cite{Feurer_2015_AAAI} and warm-start initialization~\cite{Poloczek_WSC_2016}have also been applied to hyperparameter tunning.  However,
these methods are not capable to directly
incorporate the available expert prior knowledge about the global
optimum $\pi(\boldsymbol{x}^{*})$, which is the focus of this paper.
Further, our method can also use related source observations to construct
$\pi(\boldsymbol{x}^{*})$ and thus covers the problem territory traditionally
addressed by existing transfer learning methods.

There are limited BO studies taking the prior knowledge about the optimum location into account. Siivola et al.~\cite{boundary_2017_arxiv} stated that the global
optimum is unlikely to lie in the boundary of the search space and proposed to overcome the over-exploration on boundary by placing virtual derivative
signs ('+' or '-') on the boundary. The virtual derivative signs around
the boundary are a\emph{ weak} form of prior knowledge for the GP
model so that this method does not demonstrate any advantages in experiments if the region of placing derivative signs is not large enough.  Our prior
knowledge is straightforwardly related with the location of the global optimum and is a \emph{strong} prior. Moreover, our method can support $\pi(\boldsymbol{x}^{*})$ at any location instead of only the non-boundary region assumption as~\cite{boundary_2017_arxiv,pmlr-v80-oh18a}.

Someone may argue why not reduce the search space to align with the expert prior on the global optimum. The reason is that once the expert prior is much off from the true optimum location, then the reduced search space may not contain the global optimum and one has to restart the optimization. Our proposed method can still converge with a misleading prior (refer to section \ref{subsec:Discussion}).

\section{Bayesian Optimization\label{sec:bg}}

The goal of BO is to find the global maximizer $\boldsymbol{x}_{*}=\text{argmax}_{\boldsymbol{x}\in\mathcal{X}}f(\boldsymbol{x})$
in the domain: $\mathcal{X\rightarrow\mathbb{R}}$ by using as few
evaluations as possible for an unknown and derivative-free function $f(\boldsymbol{x})$. Recall that $\mathcal{D}_{n}=\{\boldsymbol{x}_{i},y_{i}\}_{i=1}^{n}$
denotes a set of $n$ observations, where $y_{i}$ is a noisy function
value at $\boldsymbol{x}_{i}$, i.e. $y_{i}=f(\boldsymbol{x}_{i})+\varepsilon_{i}$
with $\varepsilon_{i}\sim\mathcal{N}(0,\sigma_{s}^{2})$, and $\sigma_{s}^{2}$
is a noise variance. The first step of the BO is to model the latent
function using Gaussian process (GP) \cite{Rasmussen:2005:GPM:1162254}.
A GP is a collection of random variables where the joint distribution of any finite subset of
these variables is still a Gaussian distribution. It can be specified
by the mean function $\boldsymbol{\mu}$ and the covariance function
$\mathbf{K}$. Without the loss of generality, a zero-mean
GP is often employed in BO, i.e. $f\sim\mathcal{GP}(\boldsymbol{0},\mathbf{K})$.
Then for a predicted point $\boldsymbol{x}'$, the mean and variance
of its function value can be computed as\begin{align}
 & \mu(\boldsymbol{x}')=\mathbf{k}^{T}\mathbf{K}^{-1}\boldsymbol{y}_{1:n} \nonumber\\
 & \sigma^{2}(\boldsymbol{x}')=k(\boldsymbol{x}',\boldsymbol{x}')-\mathbf{k}^{T}\mathbf{K}^{-1}\mathbf{k}
\end{align}
where $\boldsymbol{y}_{1:n}=\{y_{i}\}_{i=1}^{n}$, $\mathbf{k}=\left[k(\boldsymbol{x}',\boldsymbol{x}_{1})\,\cdots\,k(\boldsymbol{x}',\boldsymbol{x}_{n})\right]^{T}$
and the Gram matrix $\mathbf{K}=\left[k(\boldsymbol{x}_{i},\boldsymbol{x}_{j})\right]_{i,j\in\{1,\cdots,n\}}+\sigma_{s}^{2}\mathbf{I}$.
The $k$ is a kernel function and some common choices for the GP include
the square exponential (SE) kernel and the Mat\'ern kernel \cite{Rasmussen:2005:GPM:1162254}.
The second step of BO is to construct an acquisition function quantifying
the information about the next evaluation based on the GP. The popular
acquisition functions include expected improvement (EI), probability
improvement (PI) and upper confidence bound (UCB) \cite{brochu_tutorial_2010}.

Information-theoretic acquisition functions have recently emerged
since they directly measure the uncertainty about the global optimum
$\boldsymbol{x}^{*}$. Popular acquisition functions include entropy search (ES) \cite{Hennig2012_entropy}
and predictive entropy search (PES) \cite{hernandez2014predictive}, which are defined respectively
\begin{equation}
\alpha_{ES}(\boldsymbol{x})  =\mathbb{H}\big[p(\boldsymbol{x}^{*}\gv\mathcal{D}_{n})\big]  -\mathbb{E}_{p(y\gv\boldsymbol{x},\mathcal{D}_{n})}\bigg[\mathbb{H}\Big[p\big(\boldsymbol{x}^{*}\gv\mathcal{D}_{n}\cup(\boldsymbol{x},y)\big)\Big]\bigg]
\end{equation}
\begin{equation}
\alpha_{PES}(\boldsymbol{x})  =\mathbb{H}\big[p(y\gv\mathcal{D}_{n},\boldsymbol{x})\big]-\mathbb{E}_{p(\boldsymbol{x}^{*}\gv\mathcal{D}_{n})}\bigg[\mathbb{H}\Big[p(y\gv\mathcal{D}_{n},\boldsymbol{x},\boldsymbol{x}^{*})\Big]\bigg].\label{eq:PESeq}
\end{equation}
These two entropy-based methods often involve sophisticated approximations.
The common term for them is the posterior distribution $p(\boldsymbol{x}^{*}|\mathcal{D}_{n})$.
\emph{More specifically we can understand it as a conditional density
function of the random variable $\boldsymbol{x}^{*}$given $\mathcal{D}_{n}$}.
We can write it 
\begin{equation}
p(\boldsymbol{x}^{*}\gv\mathcal{D}_{n})=\int p(\boldsymbol{x}^{*}\gv f)p(f\gv\mathcal{D}_{n})df, \label{eq:Thompons}
\end{equation}
where $p(\boldsymbol{x}^{*}\gv f)=p\left(f(\boldsymbol{x}^{*})=\max{}_{\boldsymbol{x}\in\mathcal{X}}f(\boldsymbol{x})\right)$.
Eq.(\ref{eq:Thompons}) implies the generative process of $\boldsymbol{x}^{*}$
in GP model: sampling a function from the posterior $p(f\gv\mathcal{D}_{n})$
and then maximizing this function to obtain $\boldsymbol{x}^{*}$.
This process is also known as Thompson sampling (TS). For BO, at
step $i$, we randomly sample a function from the posterior GP $f_{i}\sim\mathcal{GP}$
and obtain the next evaluation $\boldsymbol{x}_{i}=\text{argmax}_{\boldsymbol{x}\in\mathcal{X}}f_{i}(\boldsymbol{x})$. The sequential TS in BO is shown in Alg \ref{alg:BOTS}.
\begin{algorithm}[t]  
\caption{Thompson Sampling for BO}
\label{alg:BOTS} 
\begin{algorithmic}[1]  
\renewcommand{\algorithmicrequire}{\textbf{Input:}}  
\REQUIRE observations $\mathcal{D}_{0}=\{\boldsymbol{x}_0,y_0\}$, the kernel $k$
\FOR {$n=1,2,\cdots$}
\STATE build the GP $\mathcal{GP}(\mu_{n},K_{n})$ conditioning on $\mathcal{D}_{n-1}$ 
\STATE randomly sample a  function $f_n\sim\mathcal{GP}(\mu_{n},K_{n})$ (section \ref{par:Samplingfunction})
\STATE obtain the next evluation $\boldsymbol{x}_n=\text{argmax}_{x\in\mathcal{X}}f_{n}(\boldsymbol{x})$;
\STATE evaluate $y_n=f(\boldsymbol{x}_n)+\varepsilon$;
\STATE augment the data $\mathcal{D}_{n}=\mathcal{D}_{n-1}\cup\{\boldsymbol{x}_n,y_{n}\}$;
\ENDFOR
\end{algorithmic}   
\end{algorithm}

\section{\label{sec:BO via PS}Bayesian optimization via Posterior Sampling\label{sec:framework}}

\begin{figure}
\begin{centering}
\subfloat[]{\begin{centering}
\includegraphics[width=0.25\columnwidth]{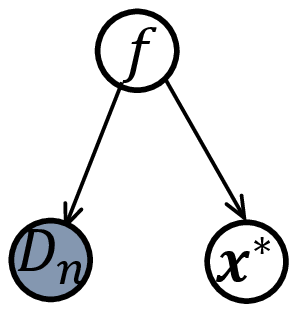}
\par\end{centering}
}\subfloat[]{\begin{centering}
\includegraphics[width=0.33\columnwidth]{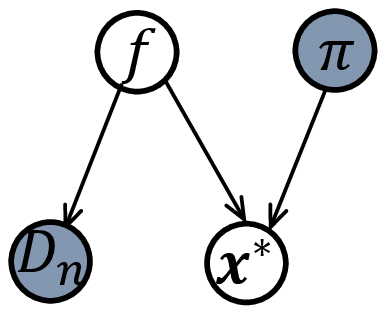}
\par\end{centering}
}
\par\end{centering}
\caption{\label{fig:The-generative-graphical}The graphical model for the optimum
$\boldsymbol{x}^{*}$ in the standard GP (a) and in the GP with the
expert prior knowledge $\pi(\boldsymbol{x}^{*})$ (b). The shadow
circle denotes observation and the plain circle denotes unobserved
variable. $\mathcal{D}_{n}$ is the function observations and $\pi$
is the expert prior knowledge about $\boldsymbol{x}^{*}$.}
\end{figure}
Suppose that the expert knowledge about the location of the global
optimum $\boldsymbol{x}_{*}$ is available in prior. Our goal is to
accelerate experimental design by incorporating this kind of expert
prior knowledge into BO. 

Recall that $\boldsymbol{x}_{*}$ is the global optimum. Since it
is unknown, we denote its estimation by $\boldsymbol{x}^{*}$. Formally,
we represent the expert prior as a probability distribution of the
random variable $\boldsymbol{x}^{*}$, denoted by $\pi(\boldsymbol{x}^{*})$
or $\pi$. We do not limit $\pi$ in non-boundary regions \cite{boundary_2017_arxiv}
while extending it to any region aligning with the global optimum.
There is no existing BO framework that can incorporate $\pi(\boldsymbol{x}^{*})$.
Our idea is to derive the posterior $p(\boldsymbol{x}^{*}\gv\mathcal{D}_{n},\pi)$
since the posterior distribution directly quantifies the uncertainty
of $\boldsymbol{x}^{*}$ and then employ posterior sampling to suggest
the next evaluation. We develop a Bayesian framework to compute the
posterior distribution of the global optimum.  

In the standard GP, the posterior distribution of $\boldsymbol{x}^{*}$
conditioning on $\mathcal{D}_{n}$ can be computed by Eq.(\ref{eq:Thompons}).
It indicates that $\boldsymbol{x}^{*}$ could be everywhere without
observations in the standard GP. The graphical model is show in Figure
\ref{fig:The-generative-graphical}(a). The process is also called
Thompson sampling. We also demonstrate the graphical model for the
GP with the expert prior $\pi(\boldsymbol{x}^{*})$ in Figure \ref{fig:The-generative-graphical}(b).
Based on the graphical model, we can infer the posterior as
\begin{align}
p(\boldsymbol{x}^{*}\gv\mathcal{D}_{n},\pi) & =\int p(\boldsymbol{x}^{*}\gv f,\pi)p(f\gv\mathcal{D}_{n})df\\
 & =\int p(\boldsymbol{x}^{*}\gv f)p(\boldsymbol{x}^{*}|\pi)p(f\gv\mathcal{D}_{n})df\\
 & \propto\pi(\boldsymbol{x}^{*})\int p(\boldsymbol{x}^{*}\gv f)p(f\gv\mathcal{D}_{n})df\label{eq:PSw1}\\
 & \propto p(\boldsymbol{x}^{*}\gv\mathcal{D}_{n})\pi(\boldsymbol{x}^{*})\label{eq:PSw2}
\end{align}
The transformation from Eq.(\ref{eq:PSw1}) to (\ref{eq:PSw2}) employs
Eq.(\ref{eq:Thompons}). 

The target posterior $p(\boldsymbol{x}^{*}\gv\mathcal{D}_{n},\pi)$,
therefore, can be inferred
\begin{equation}
p(\boldsymbol{x}^{*}\gv\mathcal{D}_{n},\pi)\propto p(\boldsymbol{x}^{*}|\mathcal{D}_{n})\pi(\boldsymbol{x}^{*})\label{eq:posterior2}
\end{equation}
Subsequently, the next evaluation point can be suggested by randomly
sampling from the target posterior above. Since the $p(\boldsymbol{x}^{*}|\mathcal{D}_{n})$
is intractable, we cannot directly sample from Eq.(\ref{eq:posterior2}).
In practice, we can sample $N$ maxima $\{\boldsymbol{x}_{i}^{*}\}_{i=1}^{N}$
from $p(\boldsymbol{x}^{*}|\mathcal{D}_{n})$ (we show how to do it
in section \ref{par:Samplingfunction}) and weight it by using $\pi$,
or
\begin{equation}
p(\boldsymbol{x}^{*}=\boldsymbol{x}_{i}^{*}\gv\mathcal{D}_{n},\pi)\propto\frac{\pi(\boldsymbol{x}_{i}^{*})}{\sum_{i=1}^{N}\pi(\boldsymbol{x}_{i}^{*}\text{)}}\label{eq:reweight}
\end{equation}
 Then a new point can be sampled from Eq.(\ref{eq:reweight}). Our
algorithm for a maximization problem is presented in Alg. \ref{alg:BO PS}.
Our proposed method incorporating prior knowledge to  update the GP posterior is straightforward to apply to PES (refer to Eq.(\ref{eq:PESeq})).
\begin{algorithm}[t] 
\caption{BO via posterior sampling (PS)}
\label{alg:BO PS} 
\begin{algorithmic}[1]  
\renewcommand{\algorithmicrequire}{\textbf{Input:}}  
\REQUIRE observations $\mathcal{D}_{0}=\{\boldsymbol{x}_0,y_0\}$, the kernel $k$, the prior distribution $\pi(x^{*})$
\FOR {$n=1,2,\cdots$}
\STATE compute $\mathcal{GP}(\mu_{n},K_{n})$ conditioning on $\mathcal{D}_{n-1}$ 
\FOR {$i=1,\cdots,N$}
\STATE randomly sample a function $f^{i}\sim\mathcal{GP}(\mu_{n},K_{n})$ and obtain the maximizer $\boldsymbol{x}^*_{i}$ (section \ref{par:Samplingfunction})
\ENDFOR
\STATE sample a new point $\boldsymbol{x}_{n}$ via Eq.(\ref{eq:reweight})
\STATE evaluate $y_n=f(\boldsymbol{x}_{n})+\varepsilon$
\STATE augment the data $\mathcal{D}_{n}=\mathcal{D}_{n-1}\cup\{\boldsymbol{x}_{n},y_{n}\}$
\ENDFOR
\end{algorithmic}   
\end{algorithm}

\subsection{\label{par:Samplingfunction}Sampling $\boldsymbol{x}^{*}$ from
posterior GP}

We can sample a function $f$ from the posterior GP and return the
global maximizer $\boldsymbol{x}^{*}$. However, evaluating such an
$f$ is very costly since it requires the complexity $\mathcal{O}(m^{3})$,
where $m$ is the number of function evaluations necessary to find
the optimum. In theory $m$ is perhaps very huge. Following the method
in \cite{hernandez2014predictive,pmlr-v70-wang17e}, we approximate
a sampled function using the linear model $f(\boldsymbol{x})\approx\boldsymbol{\phi}(\boldsymbol{x})^{T}\boldsymbol{\theta}$,
where $\boldsymbol{\phi}$ is a set of random feature and $\boldsymbol{\theta}$
is the corresponding sampled weights from its posterior distribution. 

Briefly, according to Bochner\textquoteright s theorem \cite{Bochner_fourier_1959},
a shift-invariant kernel $k$ can be rewritten as
\begin{equation}
k(\boldsymbol{x},\boldsymbol{x}^{'})=2\alpha\mathbb{E}_{p(\boldsymbol{w})}[\cos(\boldsymbol{w}^{T}\boldsymbol{x}+b)\cos(\boldsymbol{w}^{T}\boldsymbol{x}^{'}+b)]
\end{equation}
where $b\sim\mathcal{U}[0,2\pi]$, $p(\boldsymbol{w})=s(\boldsymbol{w})/\alpha$
is the normalized density of the Fourier dual $s(\boldsymbol{w})$
of the kernel $k$ and $\alpha=\int s(\boldsymbol{w})d\boldsymbol{w}$.
Further, the kernel can be approximated by $k(\boldsymbol{x},\boldsymbol{x}^{'})\approx\boldsymbol{\phi}(\boldsymbol{x})^{T}\boldsymbol{\phi}(\boldsymbol{x}^{'})$,
where $\boldsymbol{\phi}(\boldsymbol{x})=\{F^{(i)}(\boldsymbol{x})\}_{i=1}^{m}$,
$F^{(i)}(\boldsymbol{x})=\sqrt{2\alpha/m}\cos(\boldsymbol{w}_{i}^{T}\boldsymbol{x}+b_{i})$
denotes an $m$-dimensional feature mapping and $(\boldsymbol{w}_{i},b_{i})$
is a pair sampled from $p(\boldsymbol{w},b)$. For a Bayesian linear
model, conditioning on observation $\mathcal{D}_{n}$, the posterior
distribution of the weights $\boldsymbol{\theta}$ is a Gaussian distribution
$\mathcal{N}(\mathbf{m},\mathbf{v})$ with \begin{align}
 & \mathbf{m}=(\boldsymbol{\Phi}^{T}\boldsymbol{\Phi}+\sigma_{s}^{2}I)^{-1}\boldsymbol{\Phi}^{T}\boldsymbol{y}_{1:t}\\
 & \mathbf{v}=(\boldsymbol{\Phi}^{T}\boldsymbol{\Phi}+\sigma_{s}^{2}I)^{-1}\sigma_{s}^{2}
\end{align}
where $\boldsymbol{\Phi}=[\boldsymbol{\phi}(\boldsymbol{x}_{1}),\cdots,\boldsymbol{\phi}(\boldsymbol{x}_{t})]$.

\begin{figure}
\begin{centering}
\includegraphics[width=0.7\textwidth]{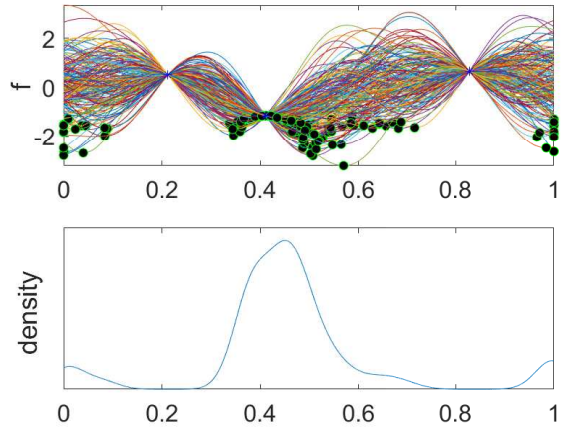}
\par\end{centering}
\caption{\label{fig:kernel-density-estimation}The top describes the Thompson
samples (colored lines) from the GP given three observations and the
returned minimizers (circles). The bottom is the density estimation
$p(\boldsymbol{x}^{*}\protect\gv\mathcal{D}_{n})$ for minimizers
shown in the top.}
\end{figure}

Let $\boldsymbol{\phi}^{(i)}(\boldsymbol{x})$ and $\boldsymbol{\theta}^{(i)}$
denote a set of random features and a sample from the posterior distribution
of $\boldsymbol{\theta}$. The sampled function can be constructed
subsequently by $f^{(i)}(\boldsymbol{x})=\boldsymbol{\phi}^{(i)}(\boldsymbol{x})^{T}\boldsymbol{\theta}^{(i)}$.
We then can maximize this function to obtain $\boldsymbol{x}_{i}^{*}=\text{argmax}_{\boldsymbol{x}\in\mathcal{X}}f^{(i)}(\boldsymbol{x})$.
We demonstrate an example for a minimization problem in Figure \ref{fig:kernel-density-estimation}.

\section{Analysis and Discussion\label{subsec:Discussion}}

In this section we would like to understand how the proposed algorithm with
a non-uniform prior on $\boldsymbol{x}^{*}$ behaves. We first make definition about an informative prior and demonstrate the robustness of our algorithm with different priors. We further provide regret insights. We denote posterior sampling
with a non-uniform $\pi$ as \textbf{PS}-G.

\begin{figure}
\begin{centering}
\includegraphics[width=0.5\columnwidth]{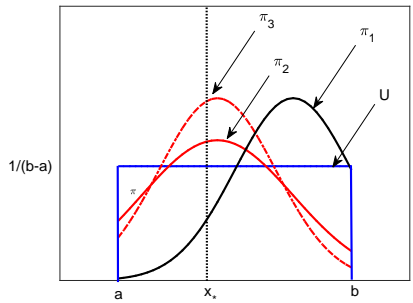}
\par\end{centering}
\caption{\label{fig:priordef}The illustration for different priors. $[a,b]$
is the support of the search space and $x_{*}$ is the true prior. `U' denotes
a uniform prior. According Definition \ref{def:def1}, $\pi_{1}$
is not an informative prior, $\pi_{2}$ and $\pi_{3}$ are informative
priors, and $\pi_{3}$ is more preferable than $\pi_{2}$. }

\end{figure}

\begin{definition}
\label{def:def1}Let the search space be $[a,b]$. The distribution
$\pi(\boldsymbol{x}^{*})$ is an informative prior if two conditions
hold: (i) the probability density of the true optimum $\boldsymbol{x}_{*}$ at
the non-uniform prior should be higher than that at the uniform prior,
i.e. $\pi(\boldsymbol{x}_{*})=\frac{1}{b-a}+r_{1}$ with $r_{1}>0$;
(ii) pick $\delta\in(0,1)$ then $\exists r_{2}$, with the probability
greater than $1-\delta$, the distance between the true optimum $x_{*}$
and the best point in the prior is smaller than $r_{2}$, i.e. $p(||\boldsymbol{x}_{*}-\boldsymbol{x}^{*}||\leq r_{2})>1-\delta$.
\end{definition}
The first condition describes the probability of sampling the true
optimum with the prior. The higher $r_{1}$ is, the more informative
$\pi$ is. The second condition describes the probability mass around
the true optimum. The lower $r_{2}$ is, the more informative $\pi$
is. We demonstrate a setting of prior in Figure \ref{fig:priordef}.

\begin{figure}
\begin{centering}
\subfloat[An example of using \emph{Thompson sampling (TS)} on a toy 1D design
problem. The top subfigure shows the the density estimation for $\boldsymbol{x}^{*}$.
$n$ is the number of observations. The bottom subfigure shows the
next evaluation suggested by TS that randomly samples a point from
the density estimation for $\boldsymbol{x}^{*}$ as the next evaluation.
The global minimizer of the synthetic function is at $\boldsymbol{x}=2$.
At $n=8$, TS almost converges to the global minimizer.]{\begin{centering}
\includegraphics[width=0.33\textwidth]{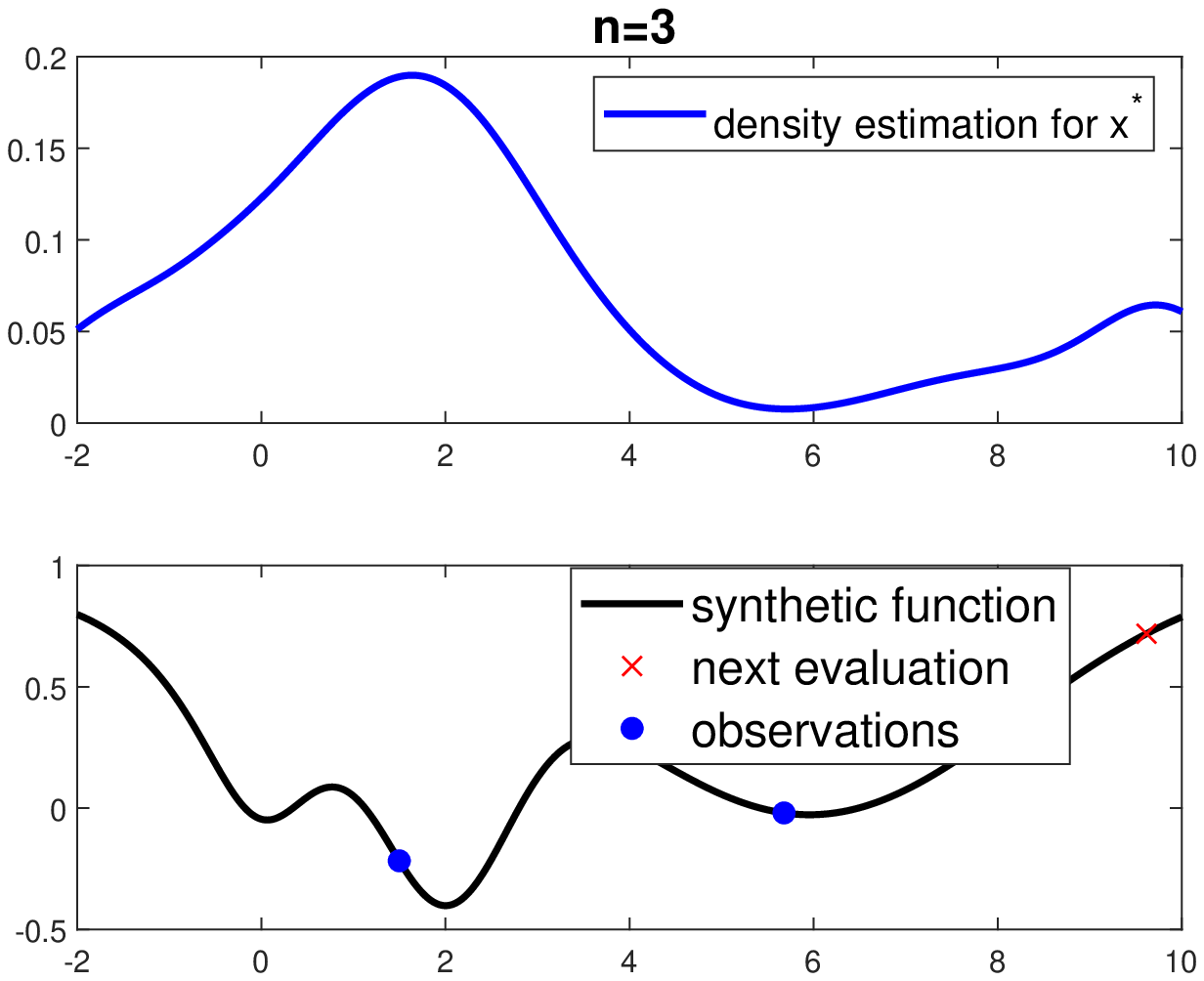}\includegraphics[width=0.33\textwidth]{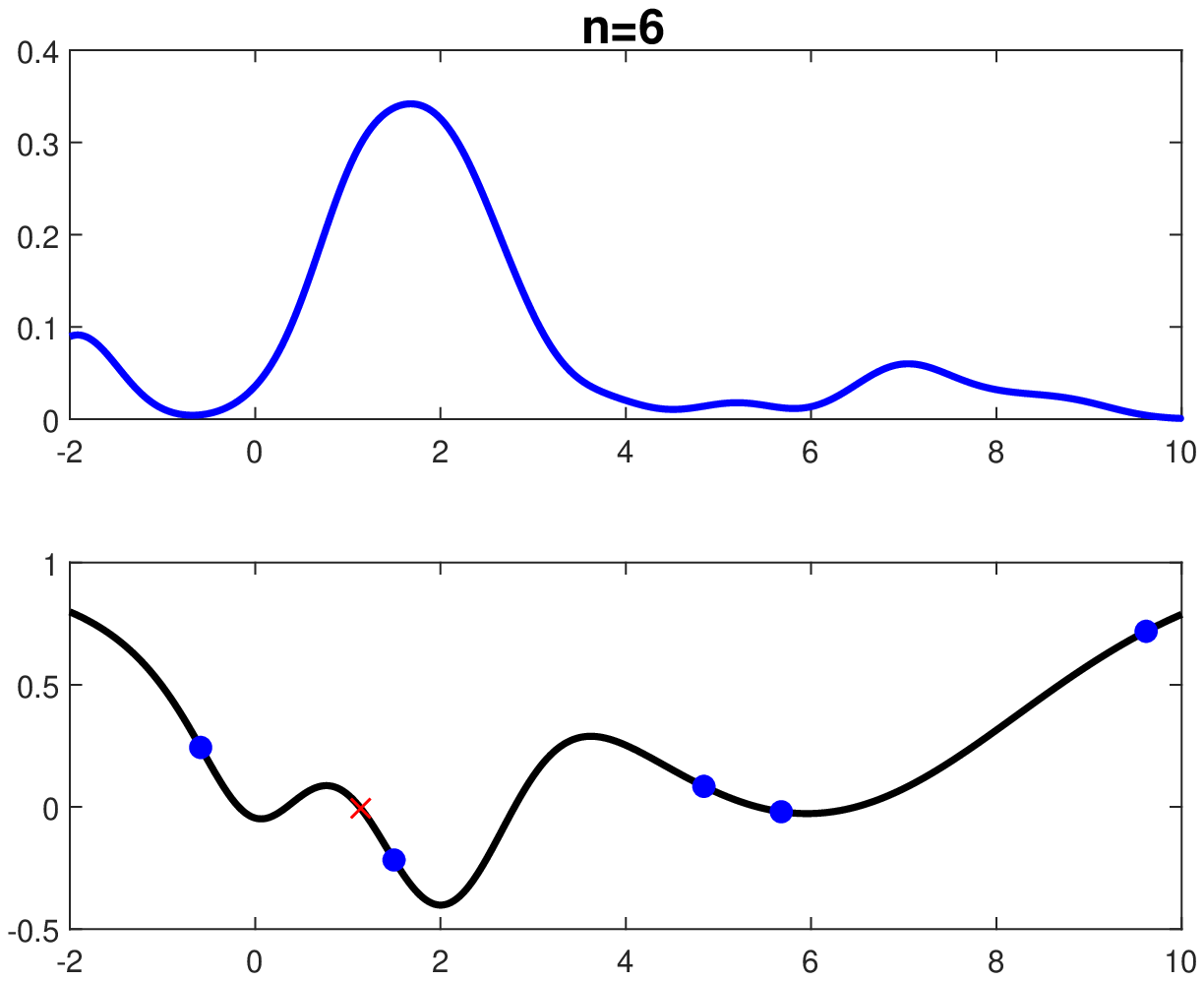}\includegraphics[width=0.33\textwidth]{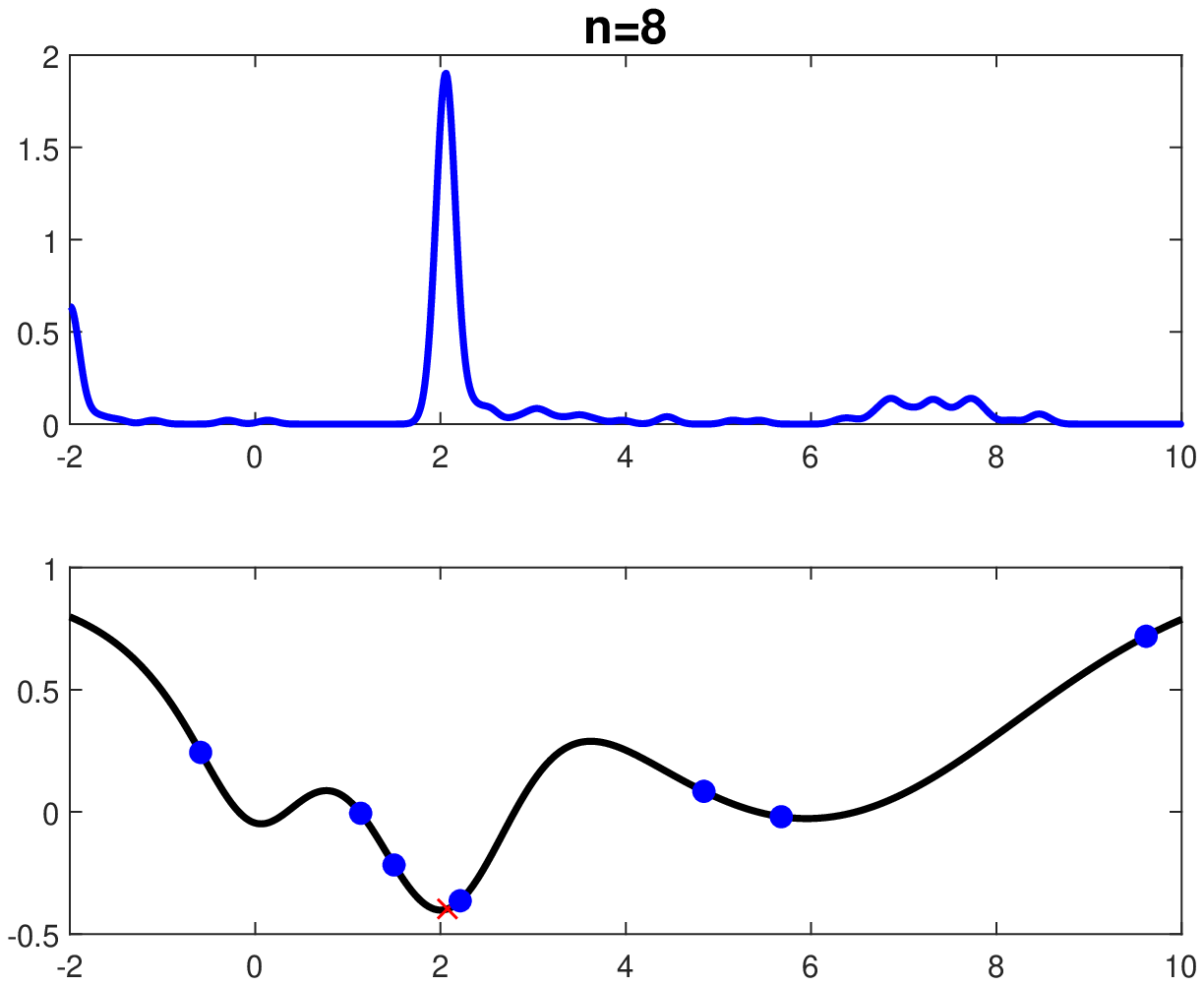}
\par\end{centering}
}
\par\end{centering}
\begin{centering}
\subfloat[An example of using \textit{posterior sampling} with a Gaussian prior $\pi(\boldsymbol{x}^{*})$
\textbf{\emph{close to}} the global minimizer. The top subfigure shows
the the density estimation for $\boldsymbol{x}^{*}$and the prior
$\pi(\boldsymbol{x}^{*})$ used in this example. The bottom subfigure
shows the next evaluation suggested by PS-G. In this case, the PS-G
sample points concentrating on the global minimizer and obtains the
narrower density estimation of $\boldsymbol{x}^{*}$.]{\begin{centering}
\includegraphics[width=0.33\textwidth]{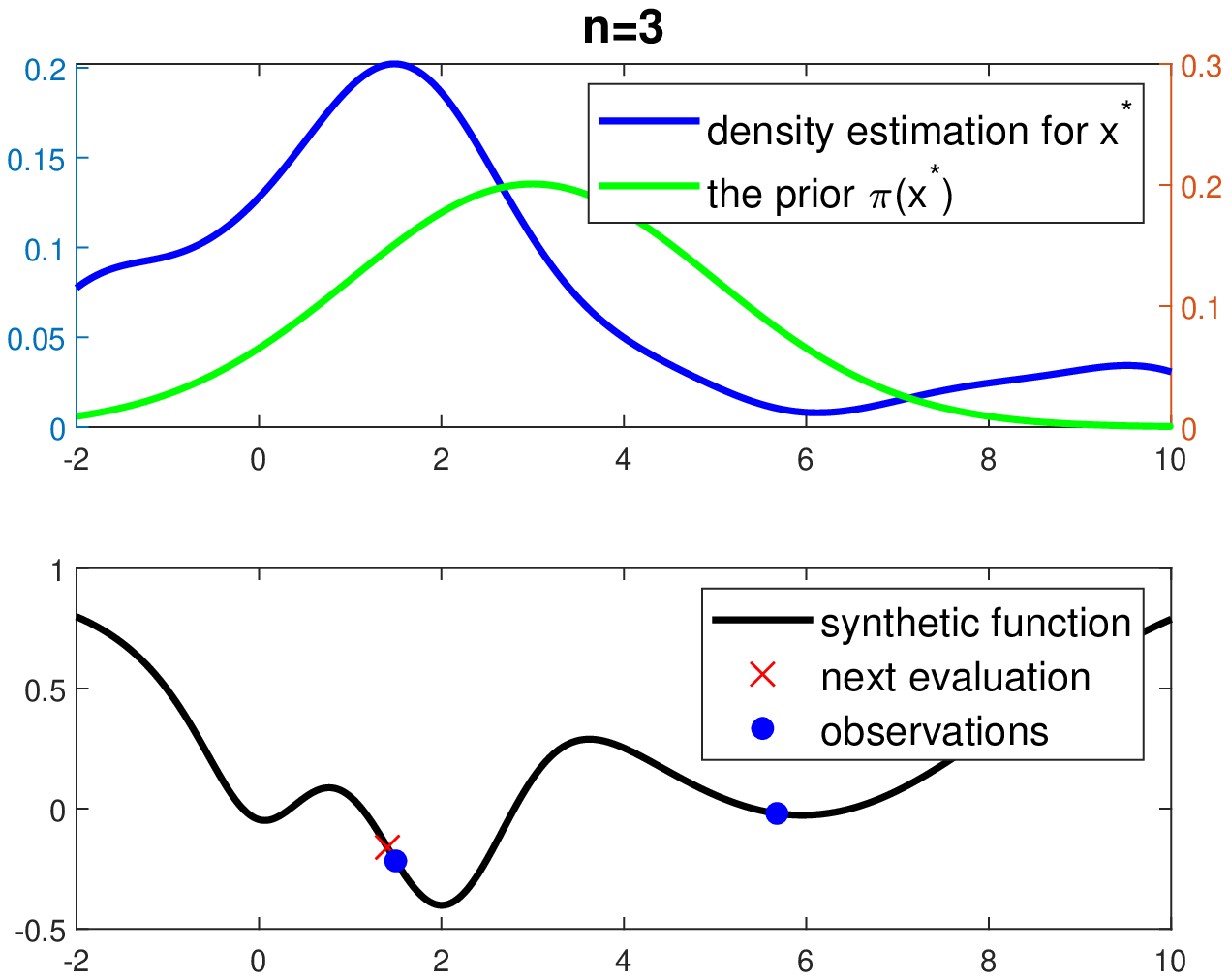}\includegraphics[width=0.33\textwidth]{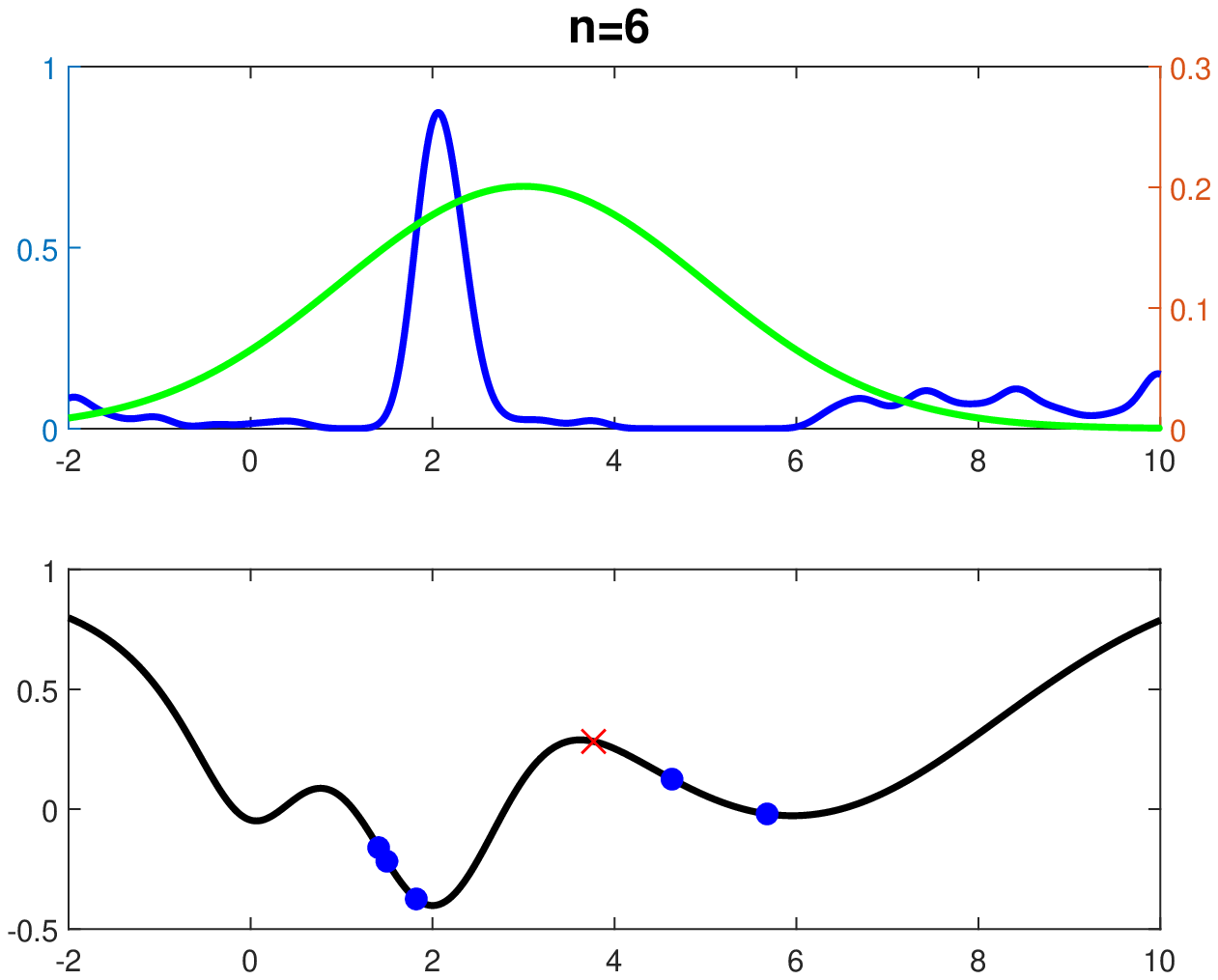}\includegraphics[width=0.33\textwidth]{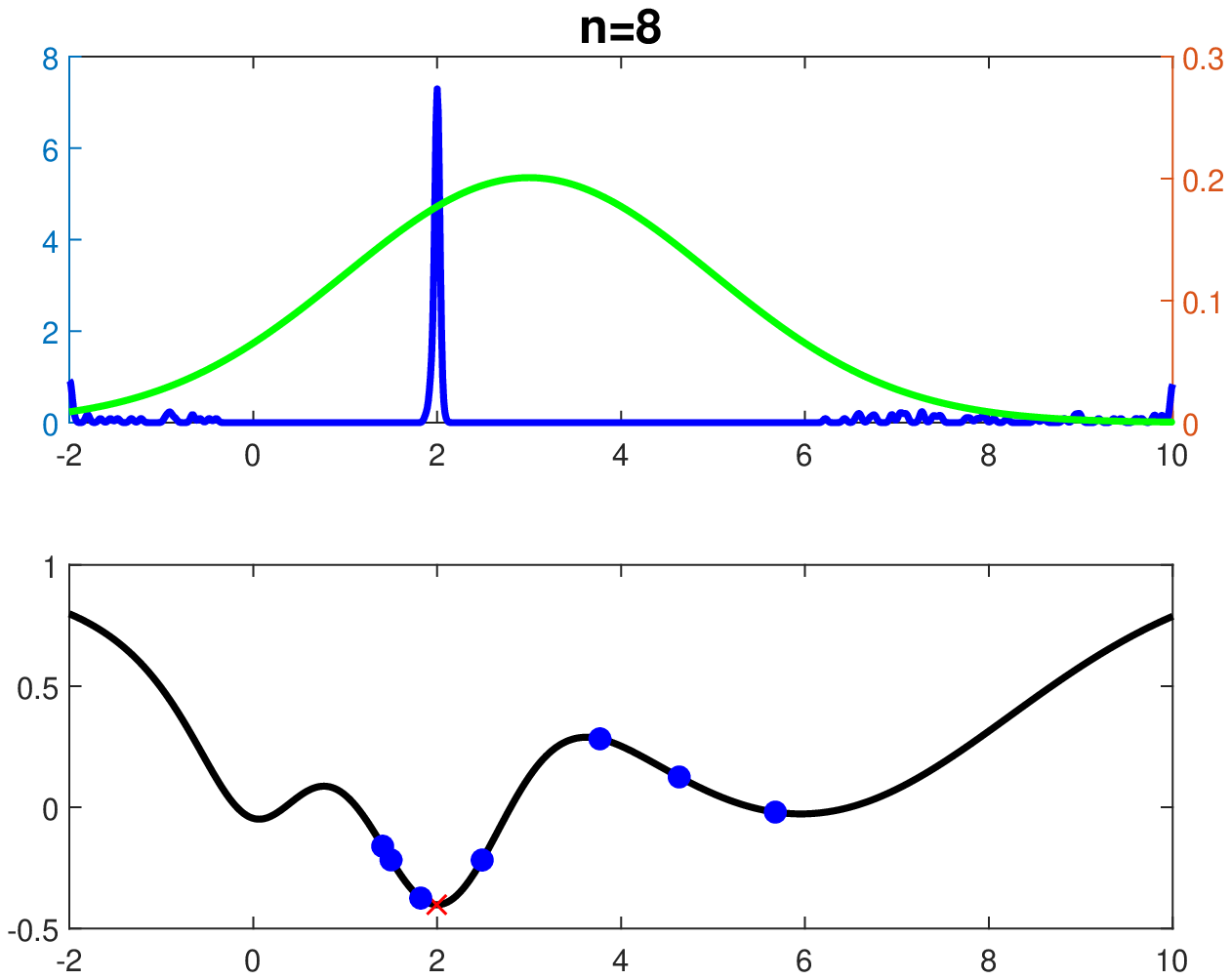}
\par\end{centering}
}
\par\end{centering}
\begin{centering}
\subfloat[An example of using \textit{posterior sampling} with a Gaussian prior $\pi(\boldsymbol{x}^{*})$
\textbf{\emph{far from}} the global minimizer. The top subfigure shows
the the density estimation for $\boldsymbol{x}^{*}$and the prior
$\pi(\boldsymbol{x}^{*})$ used in this example. The bottom subfigure
shows the next evaluation suggested by PS-G. In this case, the PS-G
sample points aligning with the prior in the beginning and then will
converge to the global minimizer since the gross prior becomes weak
compared to the the density of $\boldsymbol{x}^{*}$.]{\begin{centering}
\includegraphics[width=0.33\textwidth]{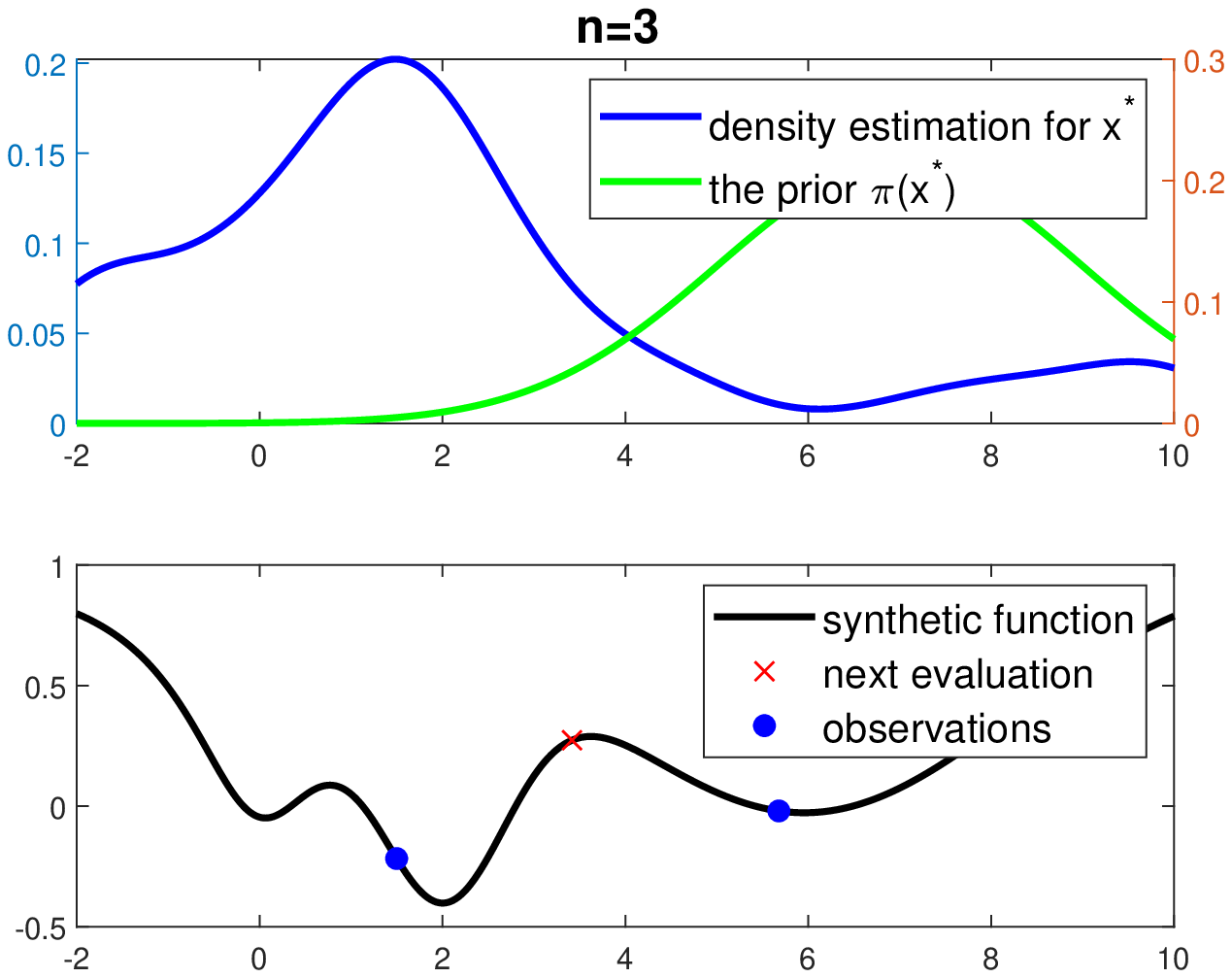}\includegraphics[width=0.33\textwidth]{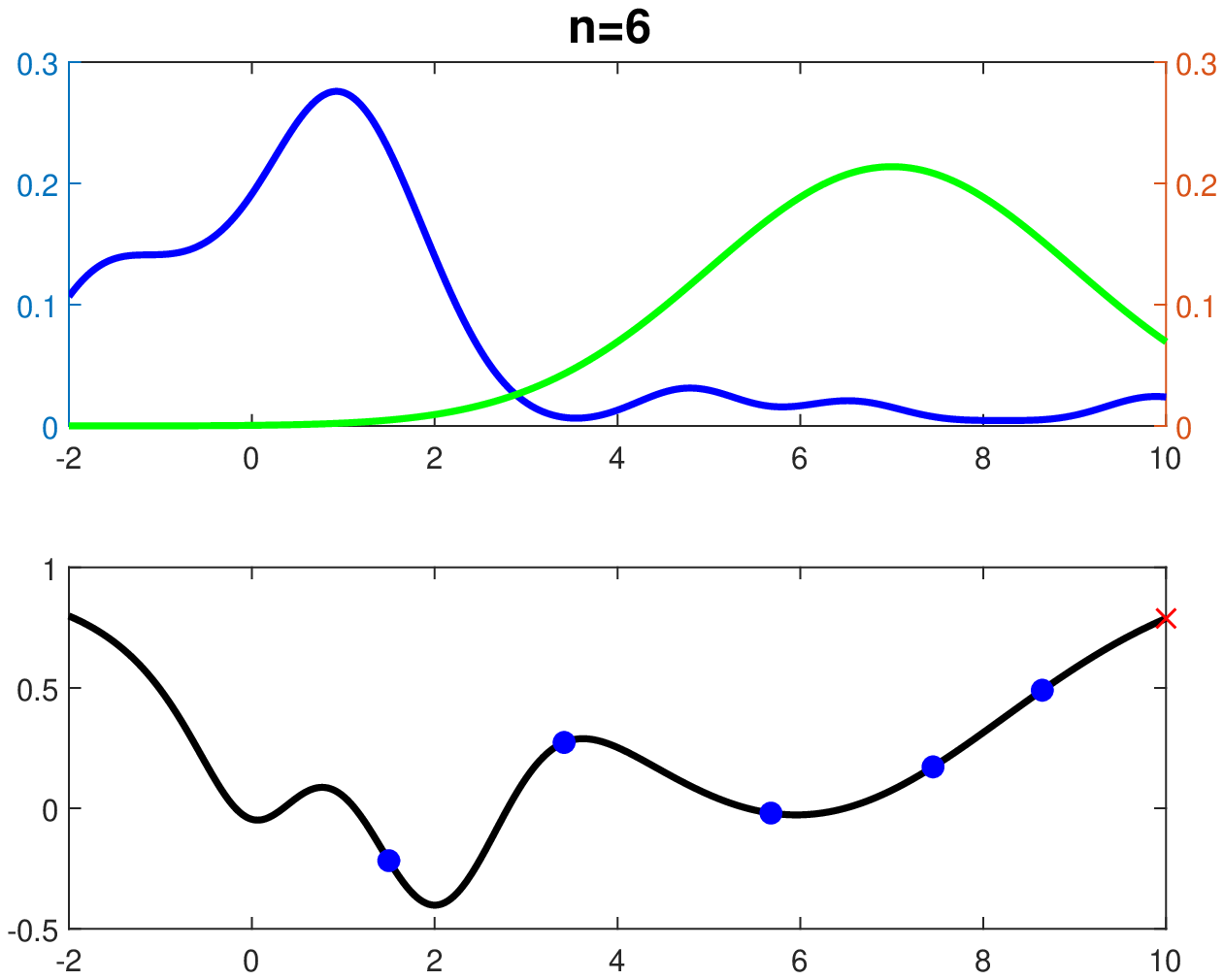}\includegraphics[width=0.33\textwidth]{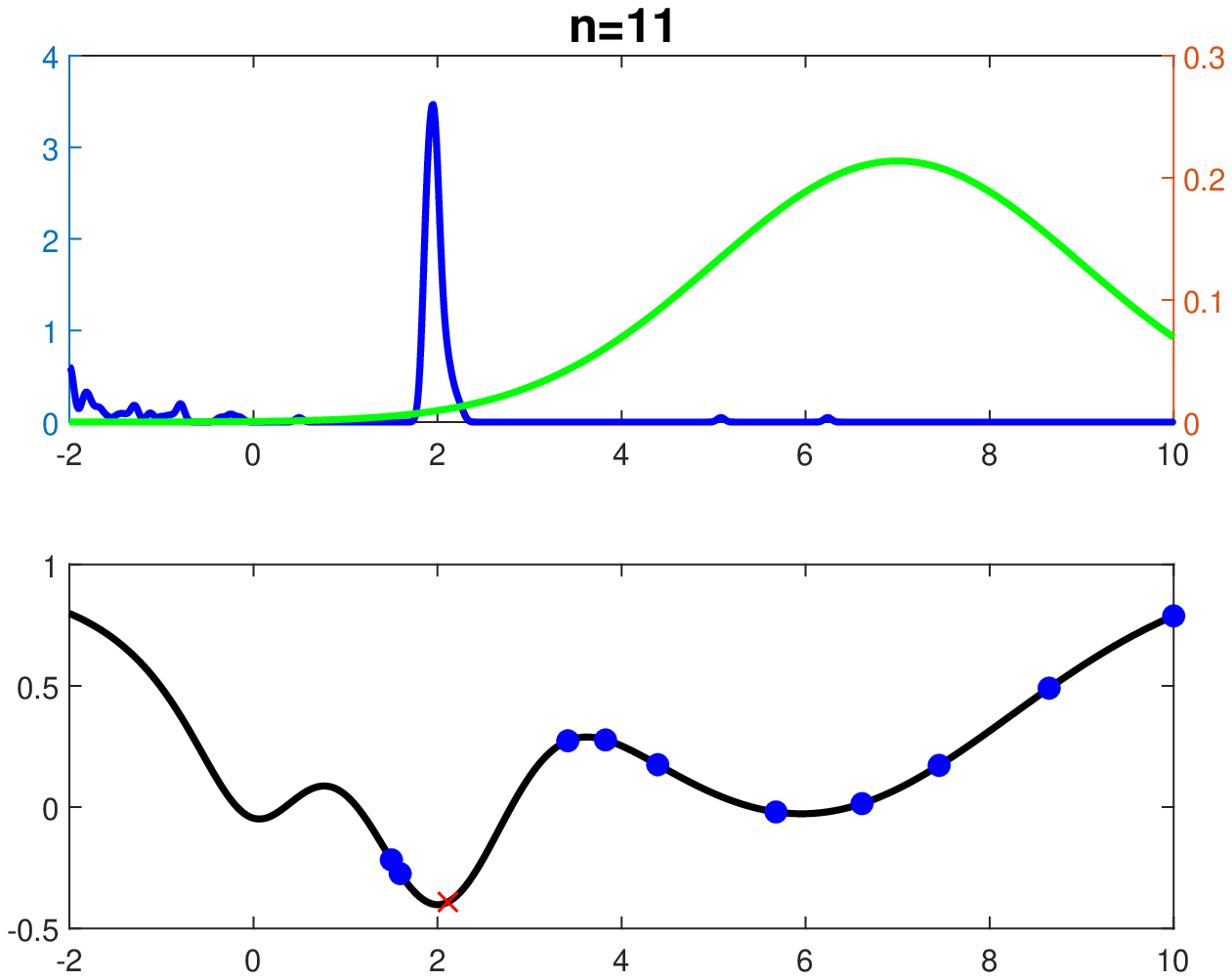}
\par\end{centering}
}
\par\end{centering}
\caption{\label{fig:TSandPS}The illustration for the robustness of PS-G on a toy 1D design problem. n is the number of iterations. The global minimizer is at $\boldsymbol{x}_{*}=2$. The top subfigures show the the density estimation for $\boldsymbol{x}^{*}$ (blue) and the prior $\pi(\boldsymbol{x}^{*})$ (green). The bottom subfigures show the next evaluation suggested by algorithms.}
\end{figure}

Based on Eq.(\ref{eq:posterior2}), we can know that PS-G with an informative prior tends to sample evaluation points concentrated around the true optimum $\boldsymbol{x}_{*}$ because the prior makes a strong impact on the GP itself posterior $p(x^{*}|\mathcal{D}_{n})$. Whilst PS-G with a non-informative (or misleading) prior first samples points by following the prior distribution. After sufficient information about $f$ has been provided by previous points, PS-G also sample near-optimum points since the GP itself posterior begins to make more effect on the combined posterior. We observe TS
and PS-G with different Gaussian priors in a 1D toy example. The results are shown in Figure \ref{fig:TSandPS}. We can see that both TS and PS-G can converge. TS tends to explore while the PS-G with a prior $\pi(\boldsymbol{x}^{*})$ close to the global minimizer tends to recommend points concentrating around the global minimizer. Note that the density estimate for $\boldsymbol{x}^{*}$ in this case is narrower than TS at the same iterations ($n=6,8$), the advantage facilitating optimization. As we expect, PS-G with a prior $\pi(\boldsymbol{x}^{*})$ far from
the global minimizer can still converge (the last figure in Figure
\ref{fig:TSandPS} (c)). It is because when we have observed sufficient data, the effect of misleading prior gets over-ridden. However, the PS-G with such a prior indeed takes more iterations for convergence. Of course, PS-G is vulnerable if with an extremely misleading prior.

Finally we analyze the regret of our algorithm with an informative prior. In BO, we are interested in the cumulative regret \cite{srinivas10gaussian}, which is defined as for a maximization problem
\begin{equation}
R(n)=\sum_{i=1}^{n}\Big[f(\boldsymbol{x}_{*})-f(\boldsymbol{x}_{i})\Bigr].
\end{equation}
Since posterior sampling uses a random policy, we compute the expected cumulative regret, called Bayesian cumulative regret, i.e.
$BR(n)=\mathbb{E}\left[R(n)\right]$. The expectation in $BR(n)$
is with respect to any randomness in the algorithm including prior
on $f$ and noisy distribution. A algorithm
is no-regret when $\lim_{n\rightarrow\infty}\frac{1}{n}BR(n)=0$.

Let $\mathcal{D}_{n}$
denote the collection of $n$ recommended points by PS-G and $\mathcal{D}_{n}^{u}$
denote the collection of $n$ recommended points by TS. ${\boldsymbol{x}}$ and $\hat{\boldsymbol{x}}$ are the realization of  $\mathcal{D}_{n}$ and $\mathcal{D}_{n}^{u}$ respectively. Then we  have $\mathbb{E}_{\hat{x}}\left[||\boldsymbol{x}_{*}-\hat{\boldsymbol{x}}||\right]\leq\mathbb{E}_{\hat{\boldsymbol{x}}^{u}}\left[||\boldsymbol{x}_{*}-\hat{\boldsymbol{x}}^{u}||\right]$
with a high probability. This is actually easy to obvious since PS-G with an informative prior can sample more concentrated points than TS we discuss before. Further, based on this equation, for a uni-modal function
$f$, we can safely conclude that the upper bound of Bayes cumulative
regret for PS-G is tighter than that of TS at
a high probability. For a multi-modal function $f$, a point closer
to the true optimum does not indicate its function value closer to $f(\boldsymbol{x}_{*})$.
Give a small distance $d$ positively relevant with $r_{2}$ in condition
(ii) of Definition \ref{def:def1}, we can know that $||\boldsymbol{x}_{*}-\hat{\boldsymbol{x}}||\leq d$
hold with a high probability. If $f$ is a Lipschitz-continuous function,
then for any point in the support of $f$ , we have $|f(\boldsymbol{x}_{*})-f(\boldsymbol{x})|\leq L||\boldsymbol{x}_{*}-\boldsymbol{x}||$.
With a further assumption $|f(\boldsymbol{x}_{*})-f(\boldsymbol{x}_{*}^{''})|\geq Ld$,
we can derive $f(\boldsymbol{x}_{*}^{''})\leq f(\boldsymbol{x}_{*})-Ld\leq f(\hat{\boldsymbol{x}})$
with a high probability. It means that PS-G still can obtain the collection of evaluation points which have higher function values than the second global optimum with a high probability. Overall, PS-G can potentially sample near-optimum evaluation points.

Similar with the regret analysis for TS in~\cite{Russo_2013_posterior,pmlr-v84-kandasamy18a}, our PS-G is also a no-regret algorithm and thus has convergence guarantee.

\section{Experiments\label{sec:experiments}}

We show the performance of our method (PS) in various tasks. We use
a truncated Gaussian  $\mathcal{N}(\boldsymbol{\mu},\boldsymbol{\Sigma})$
or Gamma prior $\Gamma(\boldsymbol{\alpha},\boldsymbol{\beta})$ because
they exhibit different notions of tail decay and are widely used
in machine learning. There does not exist a BO method which can directly
incorporate the expert prior $\pi(\boldsymbol{x}^{*})$. The algorithms
compared in this paper are:
\begin{itemize}
\item Our algorithm (\textbf{\emph{PS or PS-G or PS-G-}}\emph{$\boldsymbol{\mu}$}\textbf{\emph{-$\boldsymbol{\Sigma}$}})
\item The non-boundary search method (\textbf{\emph{DBO}}) \cite{boundary_2017_arxiv}
\item The standard BO methods without any expert prior (\textbf{\emph{TS,
PES and EI}})
\item The random search on the given prior (\textbf{\emph{Prior-based}}\textbf{
}\textbf{\emph{Random search}})
\end{itemize}
The prior-based Random search can measure the impact of $\pi(\boldsymbol{x}^{*})$.
We use the SE kernel $k(\boldsymbol{x},\boldsymbol{x}^{'})=\gamma^{2}\exp(-\Sigma_{i=1}^{D}\frac{1}{2l_{i}^{2}}(x_{i}-x_{i}^{'})^{2})$
for GP modeling, where $\gamma^{2}$ is the function variance, $l_{i}$
is the lengthscale for the $i$th dimension, $D$ is the input dimension
and we use $\sigma^{2}$ as the noise variance. For the DBO, virtual
derivative sign observations are added if the next proposed point
is within 5\% of the length of the edge of the search space to any
border. The code of our algorithm is shared in https://tini.to/PJJH.
\begin{figure*}
\begin{centering}
\subfloat[]{\begin{centering}
\includegraphics[width=0.45\columnwidth]{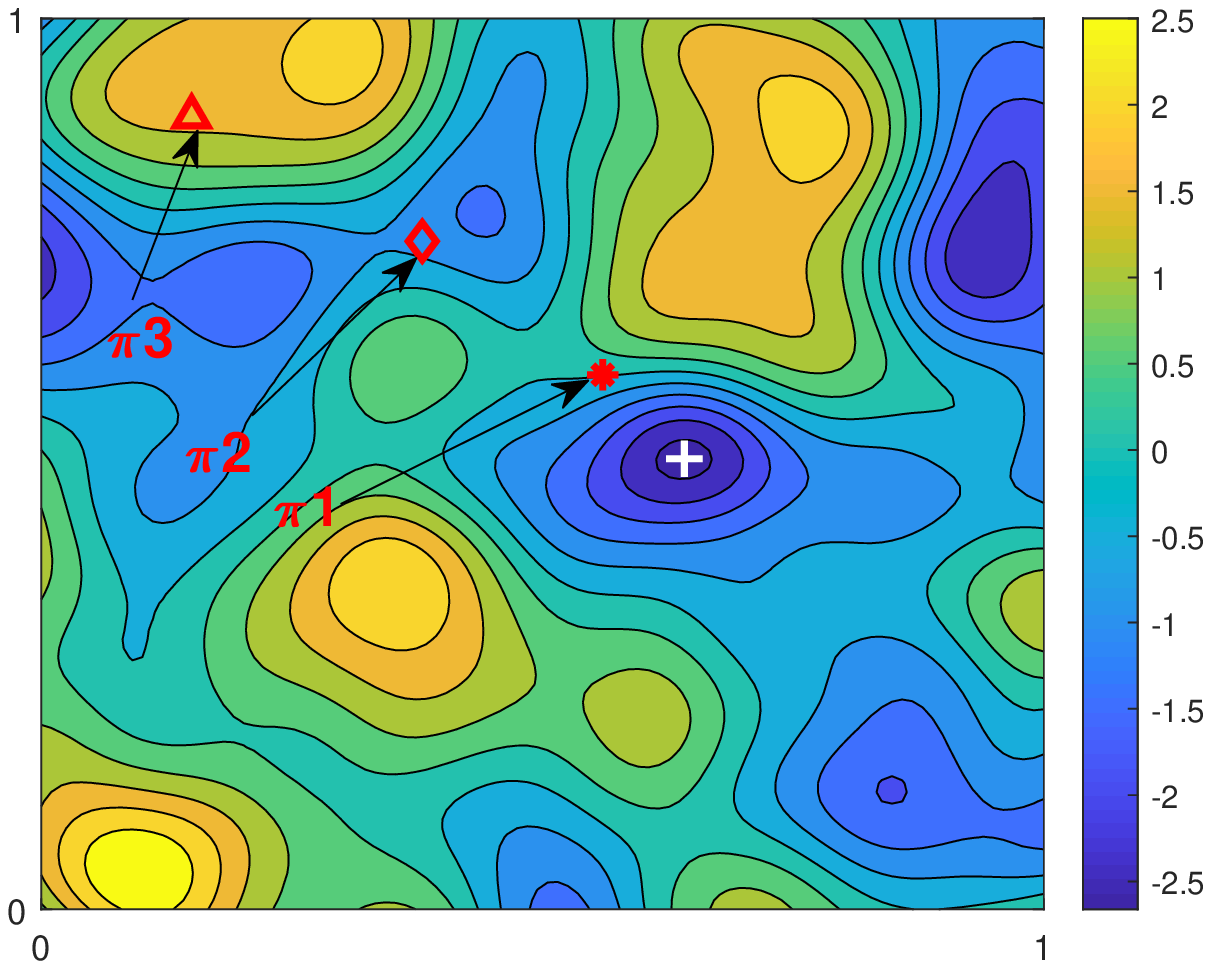}
\par\end{centering}
}
\subfloat[]{\begin{centering}
\includegraphics[width=0.45\columnwidth]{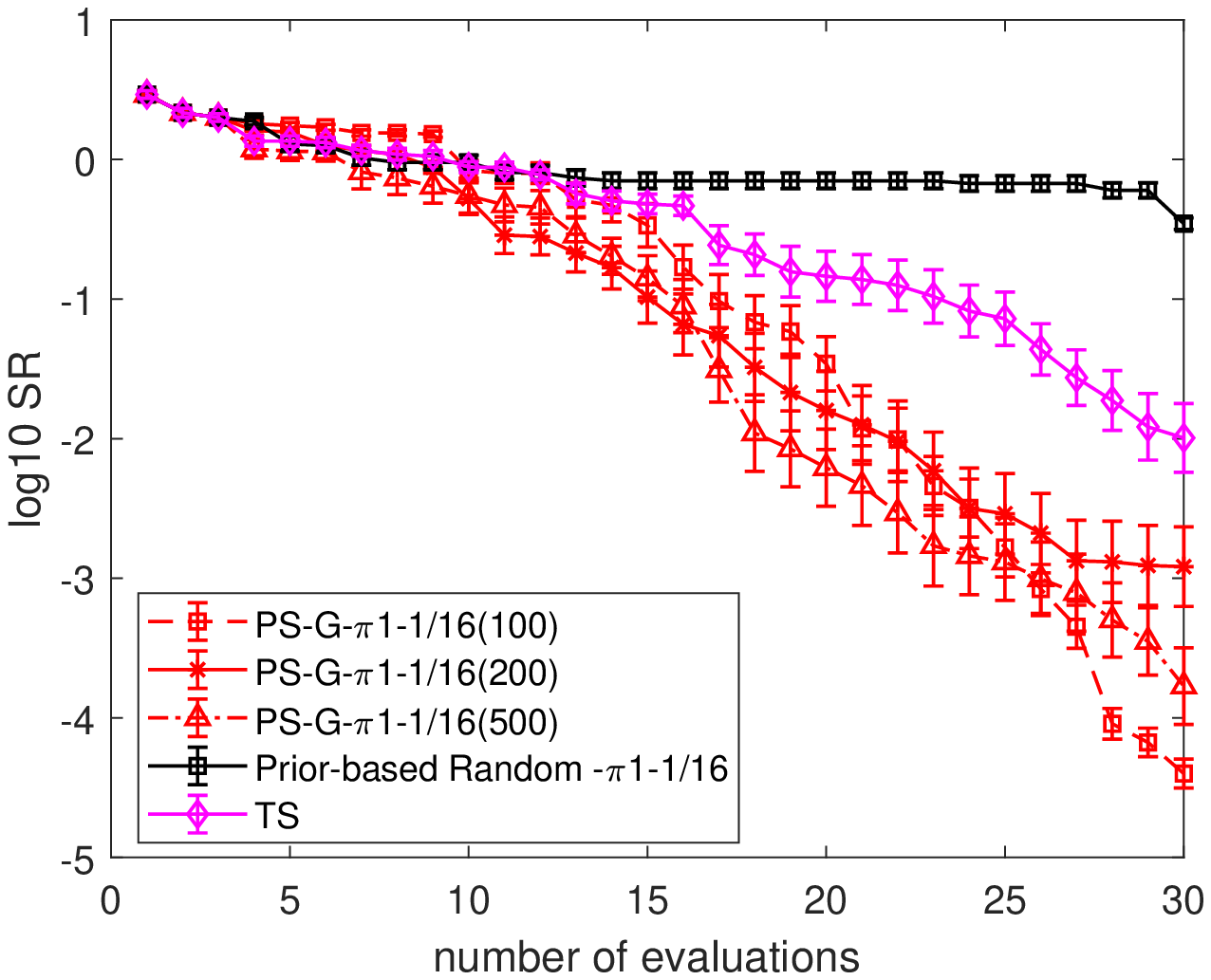}
\par\end{centering}
}
\par\end{centering}

\begin{centering}
\subfloat[]{\begin{centering}
\includegraphics[width=0.45\columnwidth]{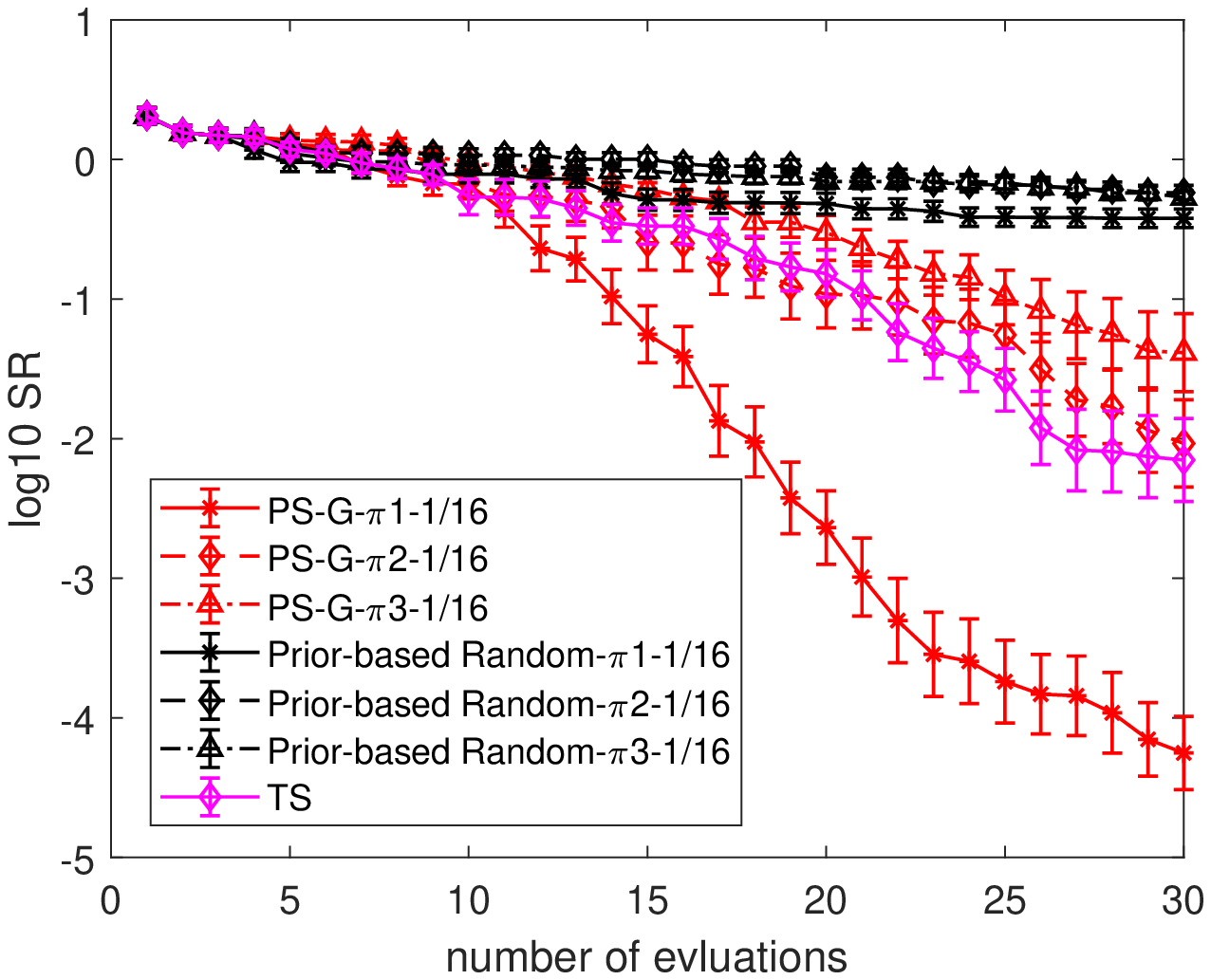}
\par\end{centering}
} \subfloat[]{\begin{centering}
\includegraphics[width=0.45\columnwidth]{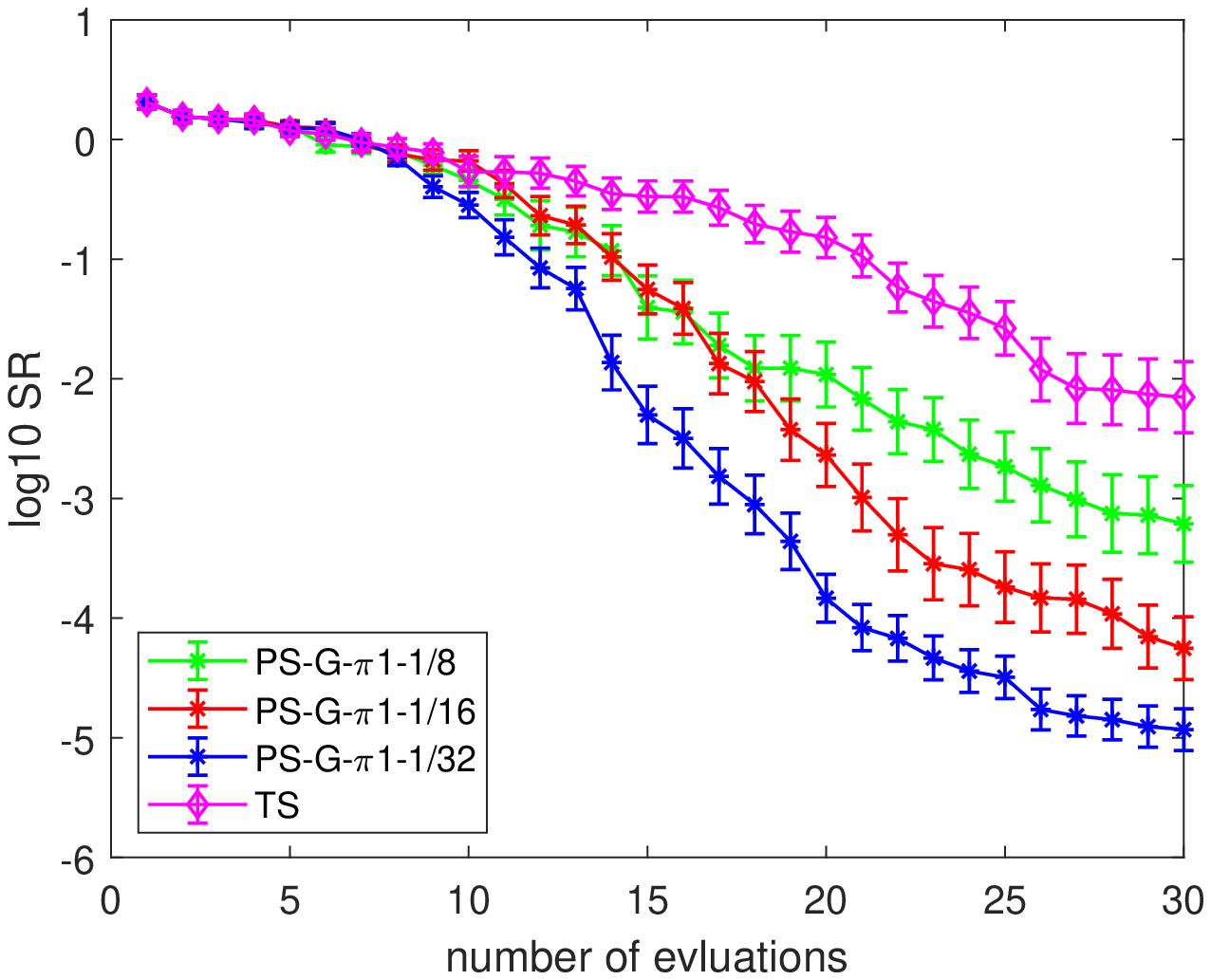}
\par\end{centering}
}
\par\end{centering}

\begin{centering}
\subfloat[]{\begin{centering}
\includegraphics[width=0.45\columnwidth]{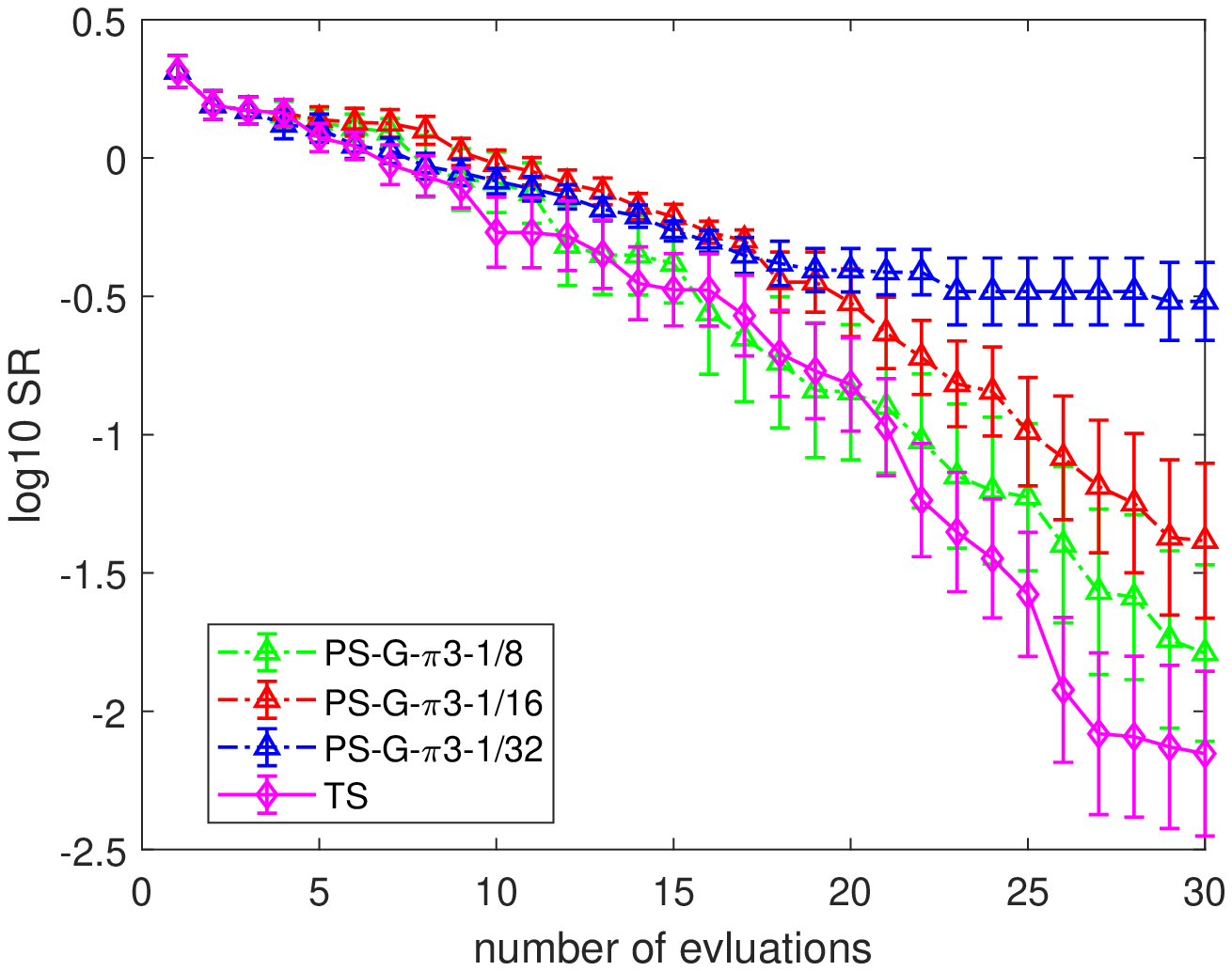}
\par\end{centering}
}\subfloat[]{\begin{centering}
\includegraphics[width=0.45\columnwidth]{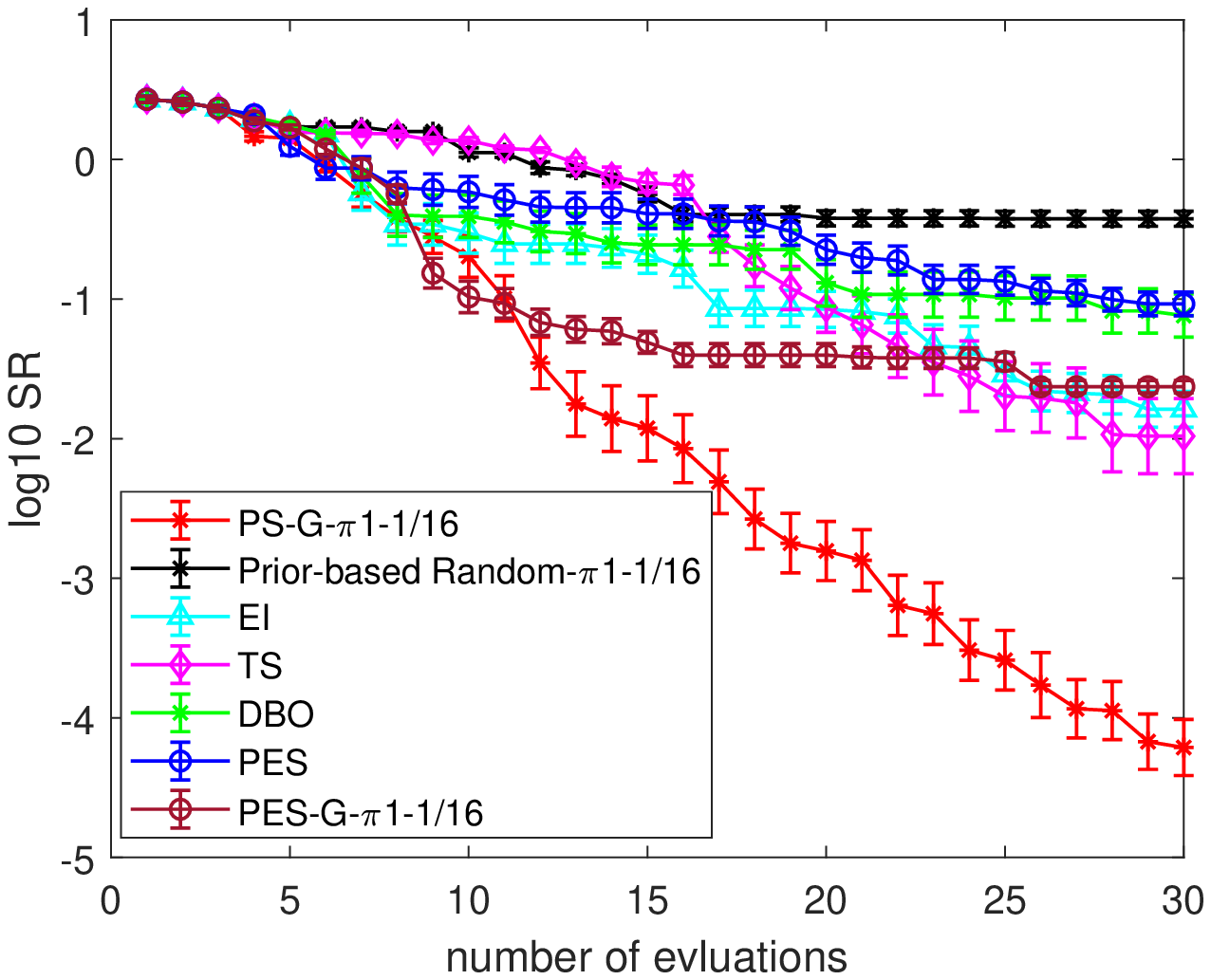}
\par\end{centering}
}
\par\end{centering}
\caption{\label{fig:objfunc}(a) The 2D objective function sampled from a known
GP prior. The white '+' denotes the global minimizer. The red symbols
denote the modes of Gaussian priors `$\pi1$', `$\pi2$' and `$\pi3$'
. (b) optimizing the 2D synthetic function by PS-G with different
number of Thompson samples. (c) optimizing the 2D synthetic function
by PS-G with the priors of different means but the same covariance
$1/16\mathbf{I}$. (d and e) optimizing the 2D synthetic function
by PS-G with the priors of the same mean ($\pi1$ and $\pi3$) but
the different covariance. (d) optimizing the 2D synthetic function
by different algorithms. }
\end{figure*}

\subsection{Optimization for Synthetic Function\label{subsec:Optimization-for-Synthetic} }

We generate a 2D objective function sampled from the GP prior with
the fixed hyperparameters $\gamma^{2}=1$, $l_{i}=0.1$ and $\sigma^{2}=10^{-6}$.
The domain is $\mathcal{X}=[0,1]^{2}$.To generate the function, we
first randomly initialize 10,000 locations in $\mathcal{X}$. Next,
we compute a covariance matrix using the GP prior mentioned above.
This covariance matrix is then used to draw a random sample for function
values from a 10,000-dimensional multi-variate normal distribution.
Then we fit a GP using these 10,000 function observations and use
the mean function of the GP as our synthetic function. We demonstrate
this function in Figure \ref{fig:objfunc} (a) and we are interested
in the global minimizer $\boldsymbol{x}_{*}=[0.64;0.50]$. For the
synthetic function optimization, we use the known hyperparameters
mentioned above into the SE kernel. We started from 3 observations
obtained by Latin hypercuber sampling in the domain and repeated experiments
20 times with different initializations. We reported the average and
standard errors of simple regret (SR).

\paragraph{On the number of Thompson samples }

Since the Eq.(\ref{eq:reweight}) in our algorithm depends on the
number of Thompson samples, we investigate how our algorithm performs
with different number of samples. We use the Gaussian prior `$\pi1$'
in Figure \ref{fig:objfunc} (a) with mean $\boldsymbol{\mu}=[0.56;0.60]$
and covariance matrix $\boldsymbol{\Sigma}=1/16\mathbf{I}$ (make
sure the search space belong to 97.5\% confidence level of the prior),
where $\mathbf{I}$ is the identity matrix. We use different number
of Thompson samples (100, 200 and 500) for PS-G. A comparison in
terms of simple regret is shown in Figure \ref{fig:objfunc}(b). The
results show that PS-G outperforms TS and is little sensitive to the
number of Thompson samples. However, we recommend to use $100\times D$
Thompson samples for reliable estimation.

\paragraph{On different priors}

We investigate the behavior of the PS-G algorithm when using different
priors on $\boldsymbol{x}^{*}$. Especially we want to understand
how our algorithm behaves with different beliefs on prior. We use
three Gaussian priors with the modes denoted as `$\pi1$', `$\pi2$'
and `$\pi3$' in Figure \ref{fig:objfunc} (a). As before, we use
200 Thompson samples in our algorithms.
\begin{itemize}
\item In the first experiment, we use the Gaussian priors with different
means and the same covariance. The results in Figure \ref{fig:objfunc}
(c) show that PS-G significantly outperforms baseline algorithms and
the PS-G with the prior `$\pi1$' that is the closest to the global
optimum performs the best.
\item In the second experiment, we use the Gaussian priors with the same
mean ($\pi_{1}$) but with the different convariance. The results
in Figure \ref{fig:objfunc} (d) indicate that the stronger covariance
belief the prior that is close to the global optimum, the better the
PS-G.
\item In the third experiment, we still use the Gaussian priors with the
same mean ($\pi_{3}$) but with the different covariance. Note that
$\pi_{3}$ is far away from the true minimizer. The results in Figure
\ref{fig:objfunc} (e) indicate that the weaker covariance belief
the prior that is far from the global optimum, the better the PS-G.
It is easy to understand since the prior will gradually approach to
the uniform distribution with the convariance becomes broad. 
\item In the last experiment, we compare our PS-G algorithm with the comprised
mean ($\pi_{1}$) and covariance ($1/16\mathbf{I}$) to baselines.
The results in Figure \ref{fig:objfunc} (f) show the advantages of
the PS-G algorithm. Note that we also compare the PES with the expert
prior on $\boldsymbol{x}^{*}$ (PES-G) with the standard PES. The
result further verifies the efficiency of our method computing the
posterior distribution of $\boldsymbol{x}^{*}$.
\end{itemize}

\paragraph{Optimizing the Hartmann 6D}

The global minimizer is $\boldsymbol{x}_{*}=[0.20169,0.150011,\\
0.476874,0.275332,0.311652,0.6573]$.
Suppose we have the Gaussian prior with the mean $\boldsymbol{\mu}=[0.3,0.3,0.6,0.4,0.4,0.75]$
and the covariance matrix $\boldsymbol{\Sigma}=1/8\mathbf{I}$. Since
PES works slowly in high dimension, we only compare PS-G with EI,
TS and prior-based Random Search. The experimental result in Figure
\ref{fig:objfunc} (d) shows the effectiveness of our method.
\begin{figure}[h]
\begin{centering}
\includegraphics[width=0.65\columnwidth]{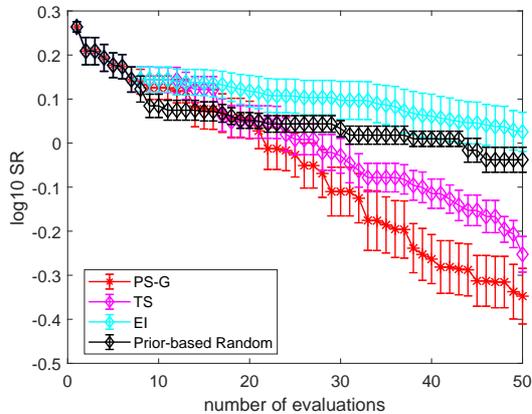}
\par\end{centering}
\caption{\label{fig:objfunc-1}simple regret on optimizing Hartmann 6D}
\end{figure}

\subsection{Hyperparameter Tuning for Classifiers}
In this experiment, we rely on a recent study by \cite{Hutter_2018_hyperparameter}
who has concluded a good range for some hyperparameters for SVM, random
forest and adaboost. For example, SVM can achieve better performance
with low values for the gamma hyperparameter and random forest tends
to perform well when using small values for minimal samples per leaf.
We were inspired by that and constructed vague hyperparameter priors
for our experiments via appropriate truncated gamma distributions.

We tune hyperparameters for three classifiers: SVM with RBF kernel
(SVM-Rbf), SVM with Sigmoid kernel (SVM-Sigmoid) and random forest
(RF). SVM-Rbf and SVM-Sigmoid use two hyperparameters - complexity
(regularization) and gamma. The range of complexity hyperparameter
is $[2^{-5},2^{15}]$ and gamma is $[2^{-15},2^{3}]$ \cite{Hutter_2018_hyperparameter}.
The RF includes two hyperparameters - `Fraction of random features
sampled per node' and `minimal samples per leaf' and the corresponding
range is $[0.1,0.9]$ and $[1,20]$. Other hyperparameters involved
with these classifiers are set to their default values as in Scikit-learn
\cite{Pedregosa_2011_scikit}. We set the noise variance $\sigma_{n}^{2}=10^{-3}$
for all algorithms and measure the average F-measure over 10-fold
cross validations. Since the optimal hyperparameters for these classifiers
are expected to locate in the boundary, the non-boundary search methods
such as D-BO \cite{boundary_2017_arxiv} and BOCK \cite{pmlr-v80-oh18a}
are not applicable.
\begin{figure*}
\begin{centering}
\subfloat[the prior for SVM]{\begin{centering}
\includegraphics[width=0.33\textwidth]{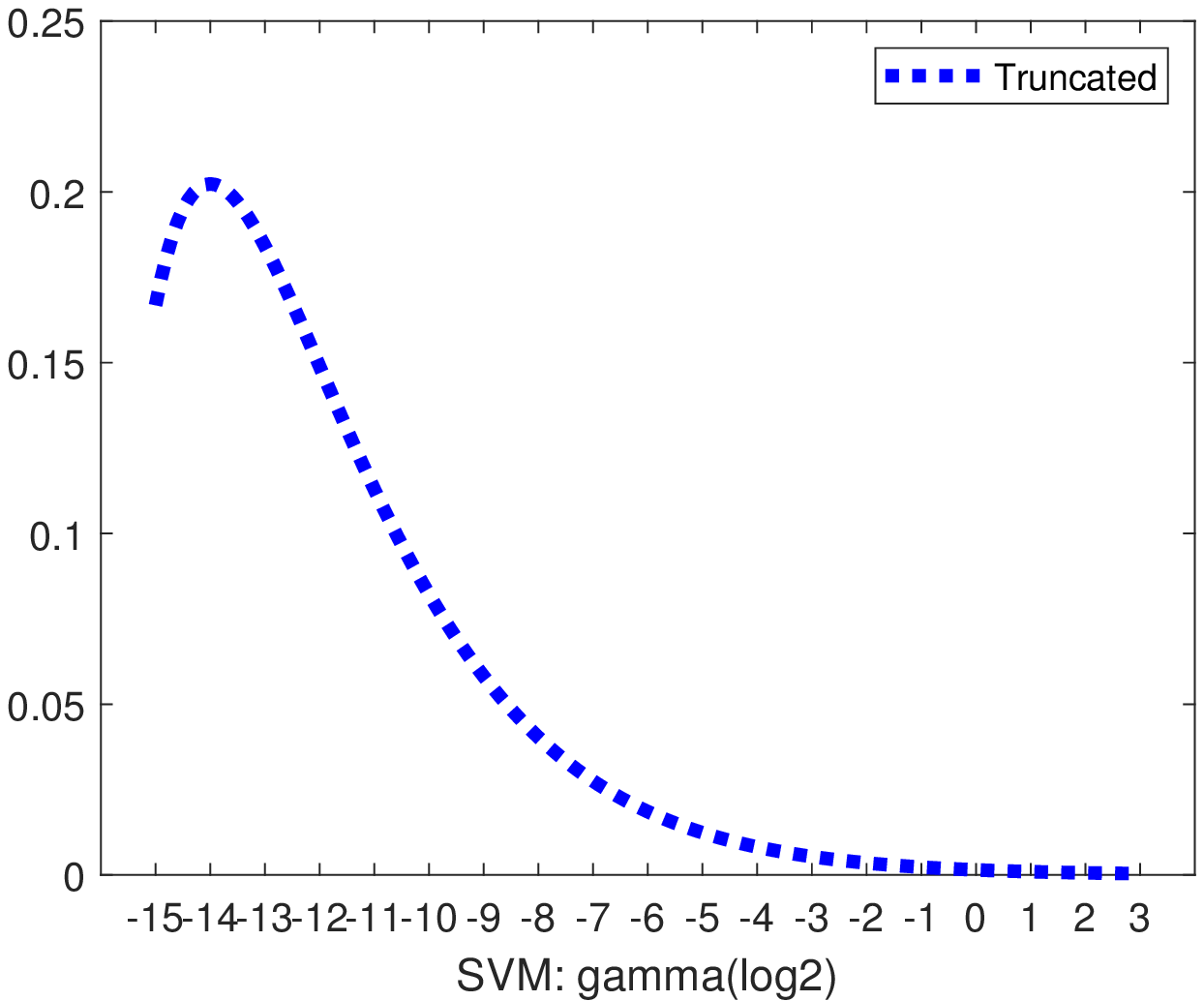}
\par\end{centering}
}\subfloat[]{\begin{centering}
\includegraphics[width=0.33\textwidth]{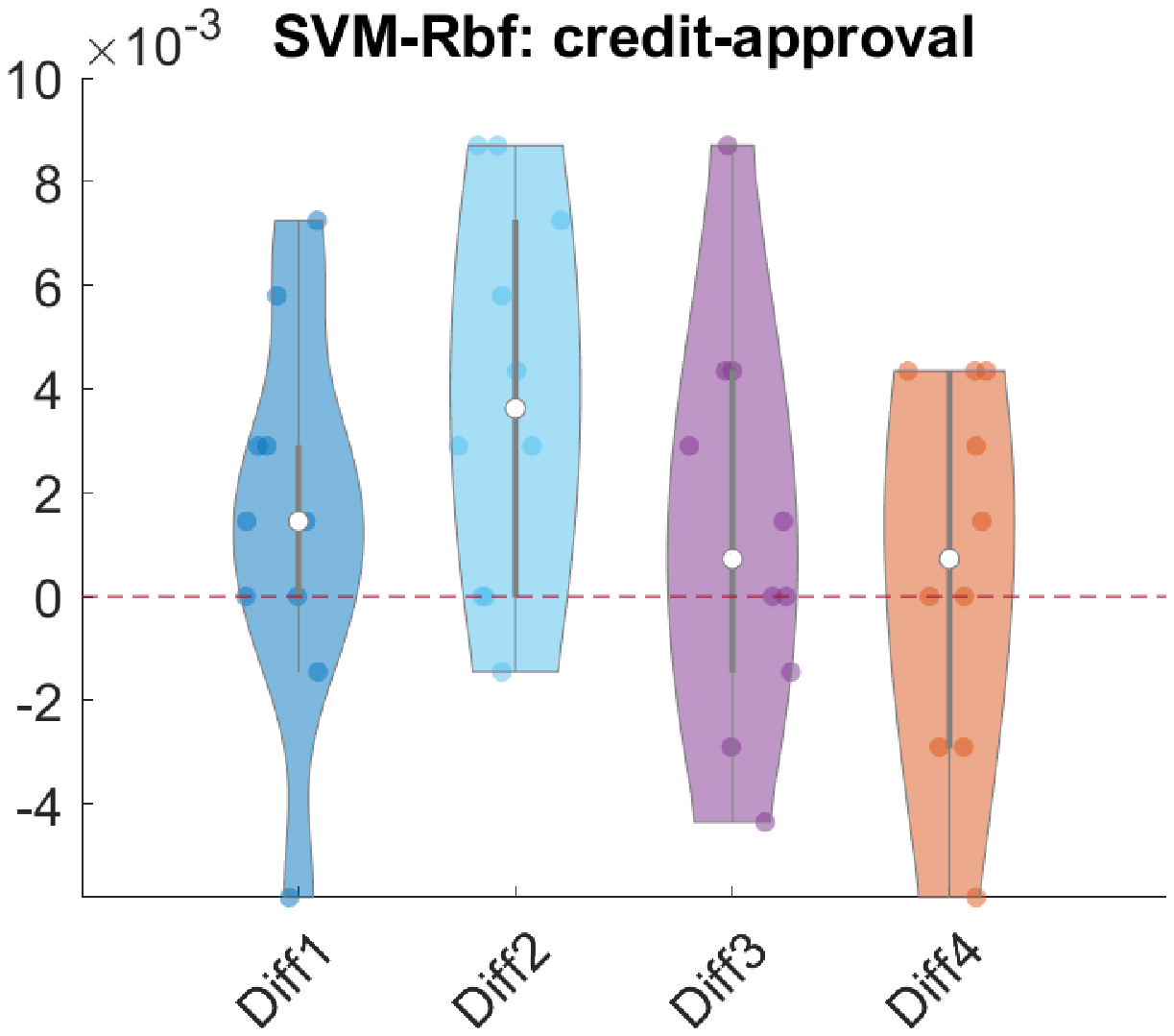}
\par\end{centering}
}\subfloat[]{\begin{centering}
\includegraphics[width=0.33\textwidth]{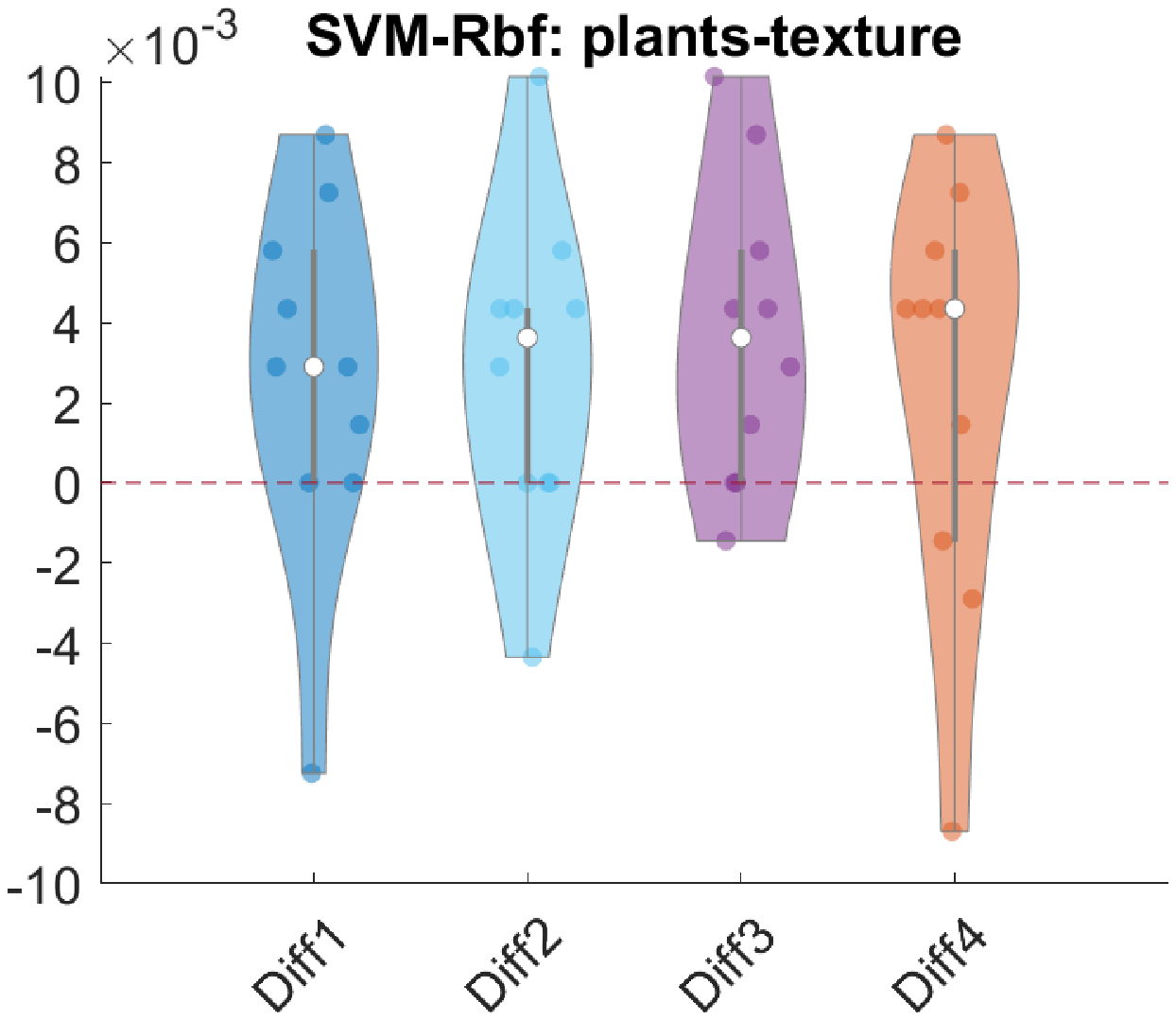}
\par\end{centering}
}
\par\end{centering}
\begin{centering}
\subfloat[]{\begin{centering}
\includegraphics[width=0.33\textwidth]{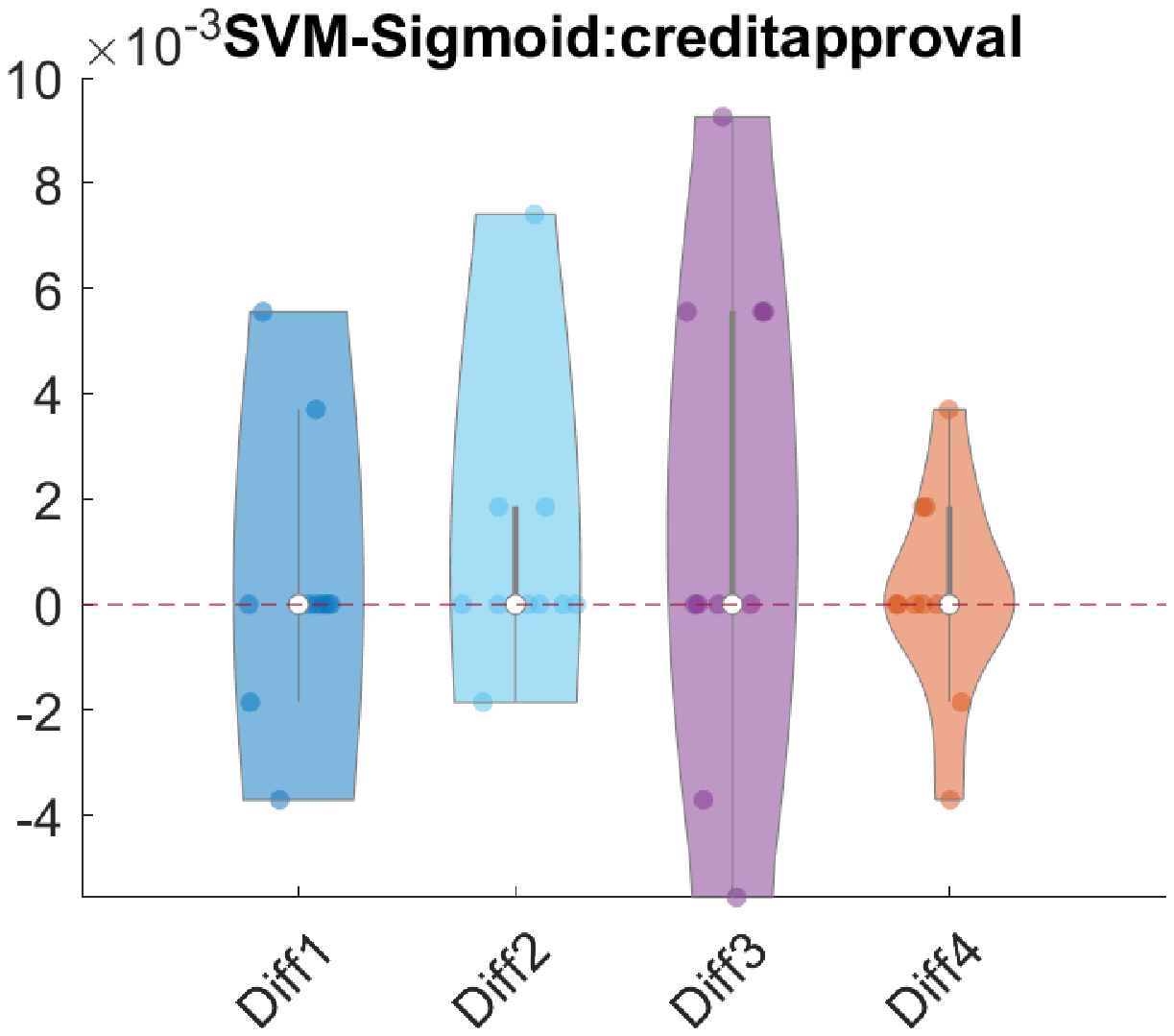}
\par\end{centering}
}
\subfloat[]{\begin{centering}
\includegraphics[width=0.33\textwidth]{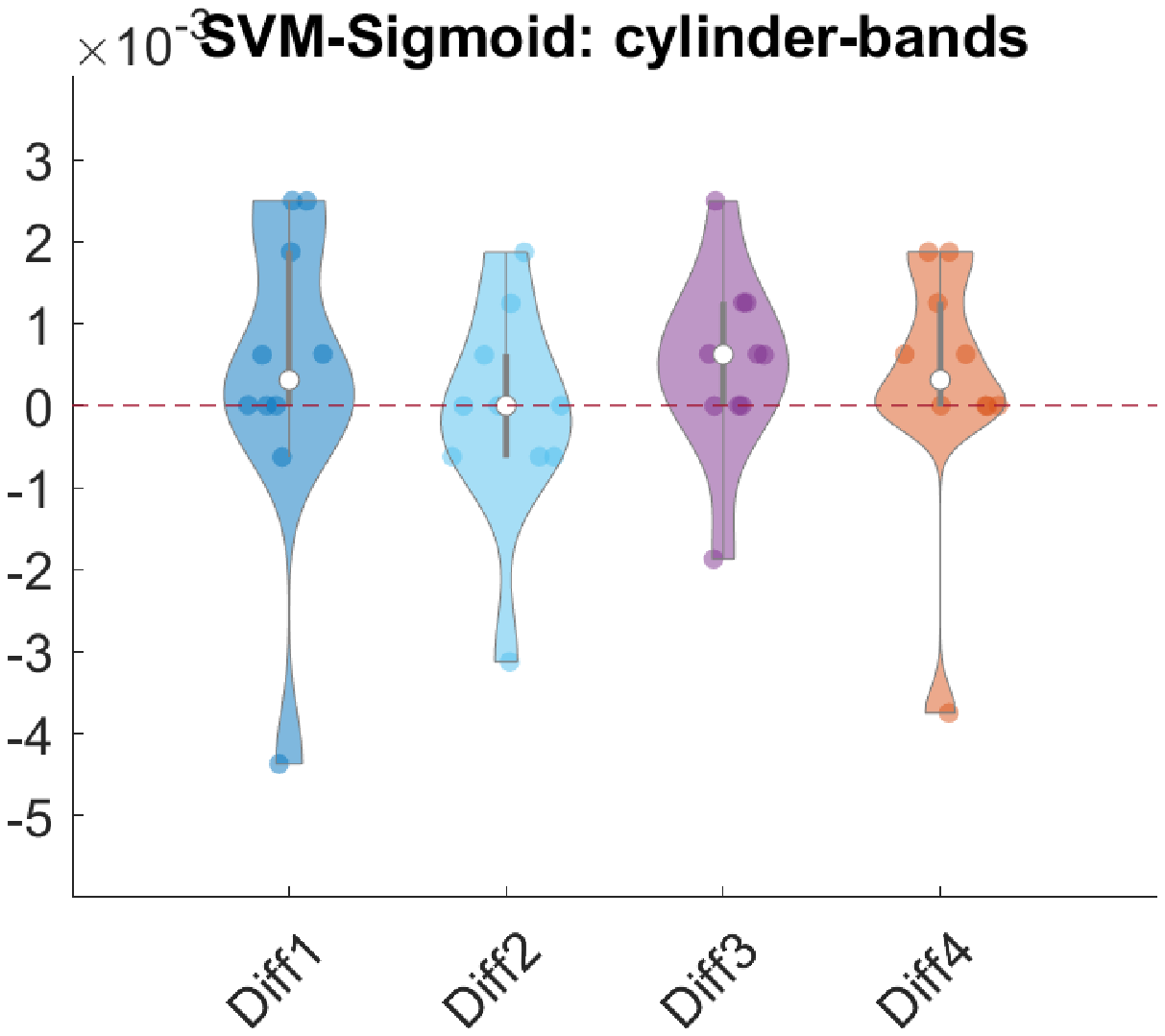}
\par\end{centering}
}
\par\end{centering}

\begin{centering}
\subfloat[the prior for RF]{\begin{centering}
\includegraphics[width=0.33\textwidth]{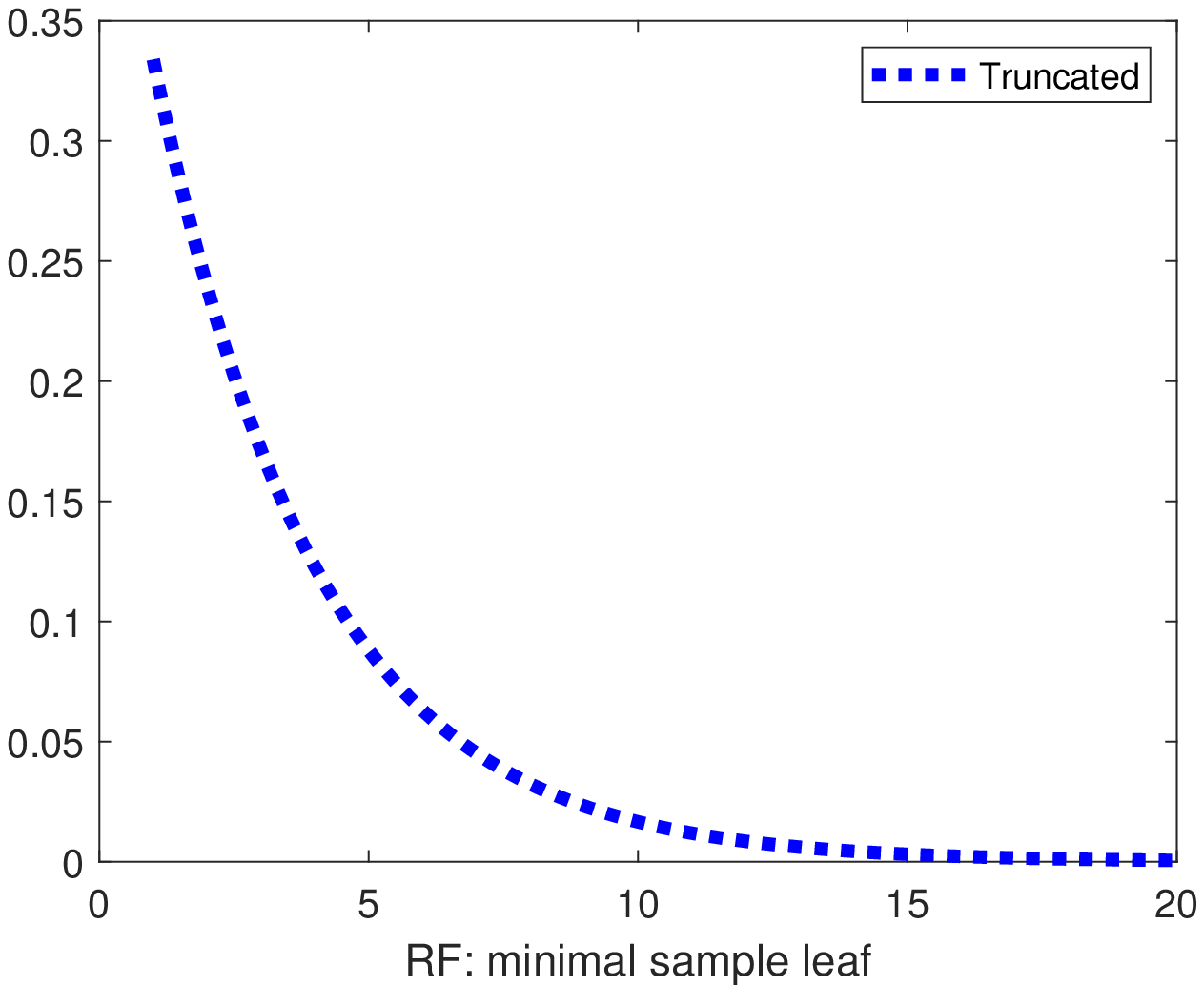}
\par\end{centering}
}\subfloat[]{\begin{centering}
\includegraphics[width=0.33\textwidth]{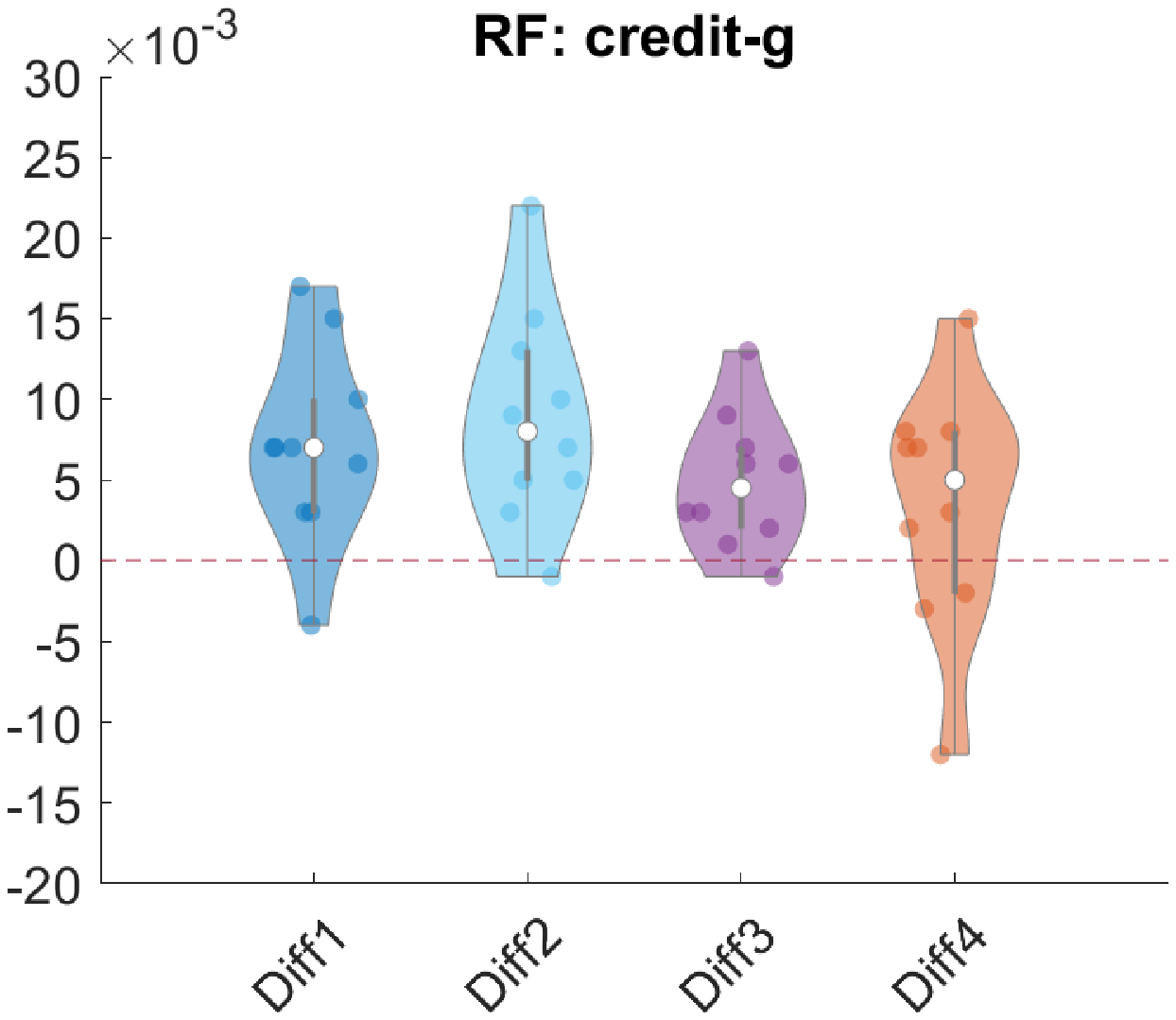}
\par\end{centering}
}\subfloat[]{\begin{centering}
\includegraphics[width=0.33\textwidth]{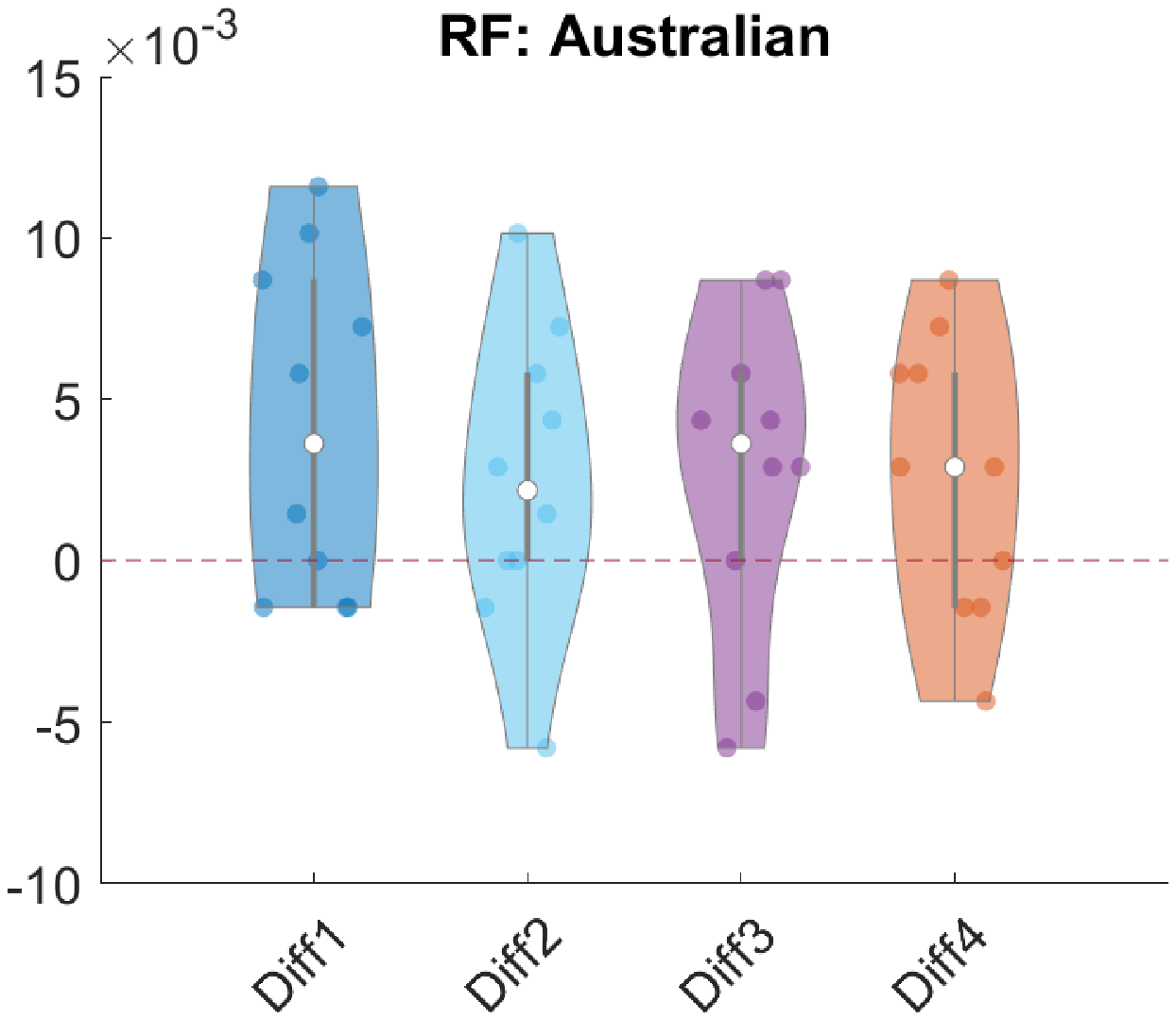}
\par\end{centering}
}
\par\end{centering}
\caption{\label{fig:hypertune} Hyperparameter tuning for SVM and Random forest.
(a): the prior $\Gamma(2,2)$ used on the hyperparameter gamma (log-scale)
in SVM-Rbf and SVM-Sigmoid. (b,c): the differences in maximal validation
accuracy during 30 iterations for SVM-Rbf between PS-G and baselines.
\textbf{'Diff1'} denotes the comparison between \textbf{PS-G and PES},
\textbf{'Diff2'} denotes the comparison between \textbf{PS-G and TS},
\textbf{'Diff3'} denotes the comparison between \textbf{PS-G and EI},
and \textbf{'Diff4'} denotes the comparison between \textbf{PS-G and
prior-based random search}. (d,e): the differences in maximal validation
accuracy for SVM-Sigmoid between PS-G and baselines. (f): the prior
$\Gamma(1,3)$ used on the minimal sample per leaf in Random forest.
(g,h): the differences in maximal validation accuracy for RF between
PS-G and baselines.}
\end{figure*}
We used the datasets from openML100 \cite{Bischl_2017_openML100}
- a comprehensive benchmark suites of machine learning datasets. The
prior distribution of hyperparameter gamma for both SVM-Rbf and SVM-Sigmoid
is a truncated Gamma distribution $\Gamma(2,2)$, shown in Figure
\ref{fig:hypertune} (a), and the prior of other hyperparameters is
a uniform distribution by default. We run the experiments for 10 times.
Figure \ref{fig:hypertune} (b,c) shows the differences in maximal
validation accuracy during 30 iterations for SVM-Rbf between the PS-G
and baselines: values greater than 0 indicate that sampling according
to PS-G was better by this amount than the baseline, and vice versa.
These differences are aggregated using a violinplot. We can see that
PS-G with the Gamma prior works best in the experiments of SVM-Rbf.
For the SVM-Sigmoid, PS-G again outperforms the baselines for the
dataset credit-approval while all algorithms are close for the cylinder-bands.

For random forest, we used the truncated Gamma prior distribution
on `minimal sample per leaf' $\Gamma(1,3)$ in Figure \ref{fig:hypertune}
(f) . The experimental results in Figure \ref{fig:hypertune} (g,h)
show the efficiency of the PS-G algorithm. 
\begin{figure*}
\centering{}\subfloat[]{\begin{centering}
\includegraphics[width=0.35\columnwidth,height=3.0cm]{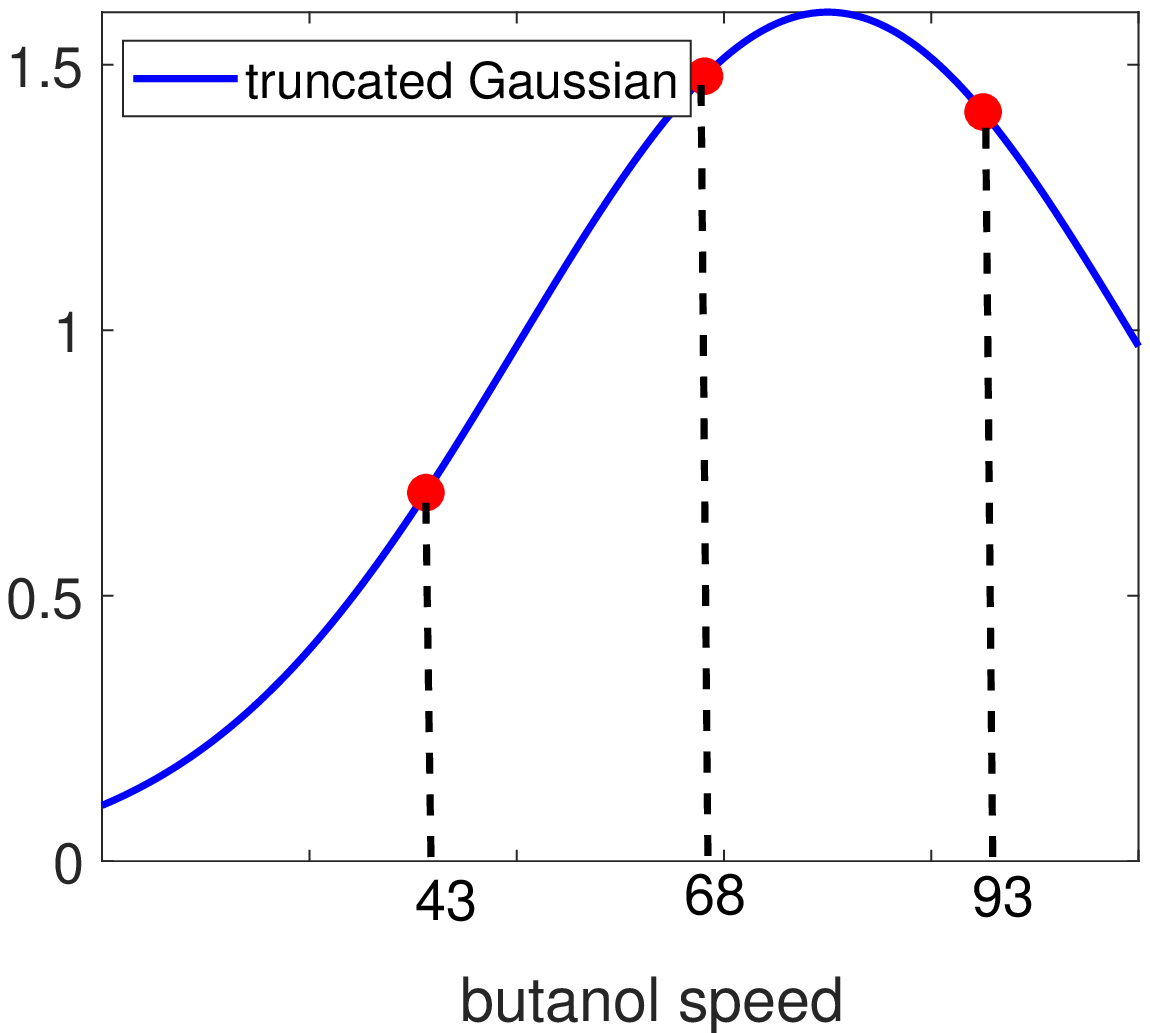}
\par\end{centering}
}\subfloat[]{\centering{}\includegraphics[width=0.35\columnwidth,keepaspectratio]{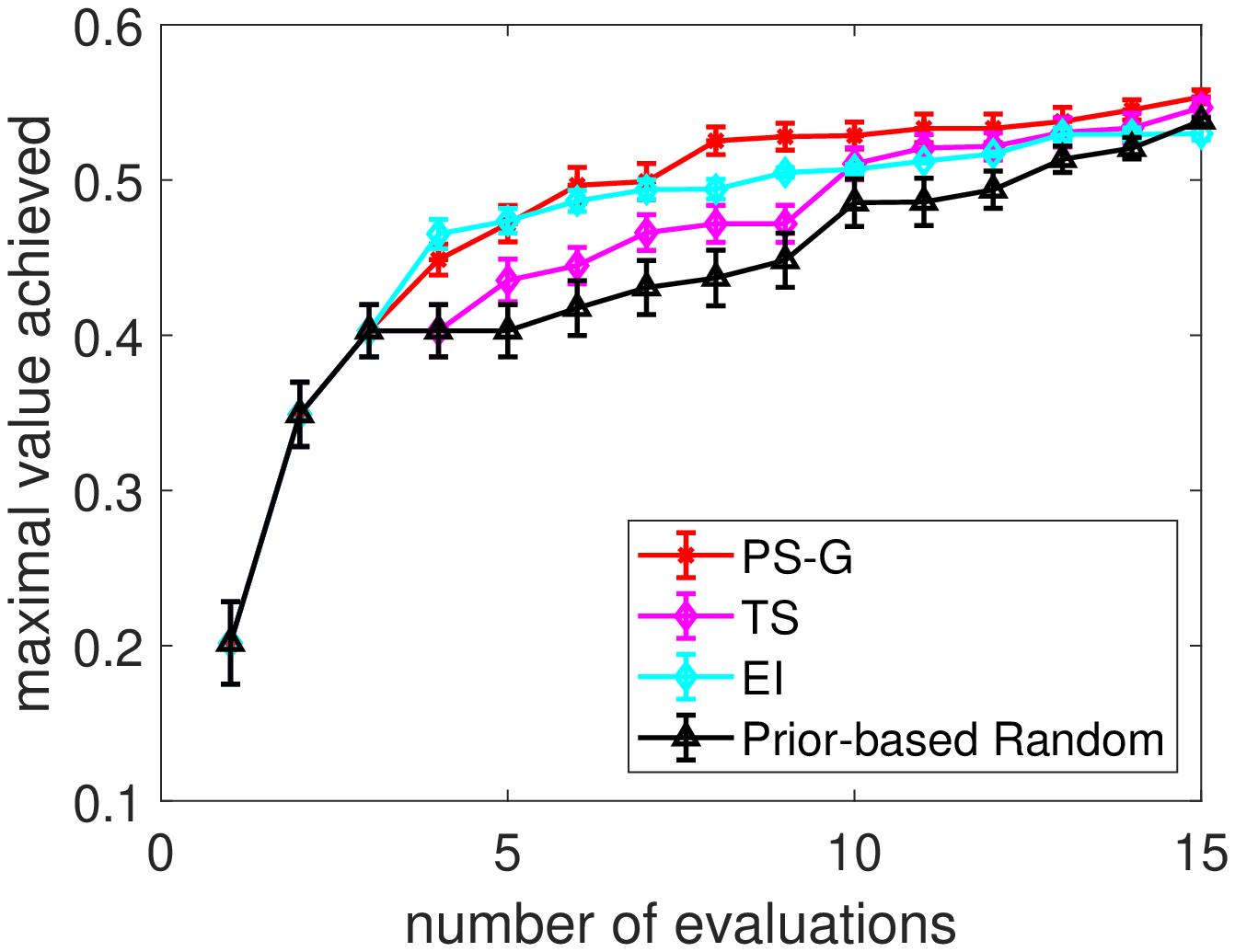}}\subfloat[]{\begin{centering}
\includegraphics[width=0.25\columnwidth,height=3.0cm]{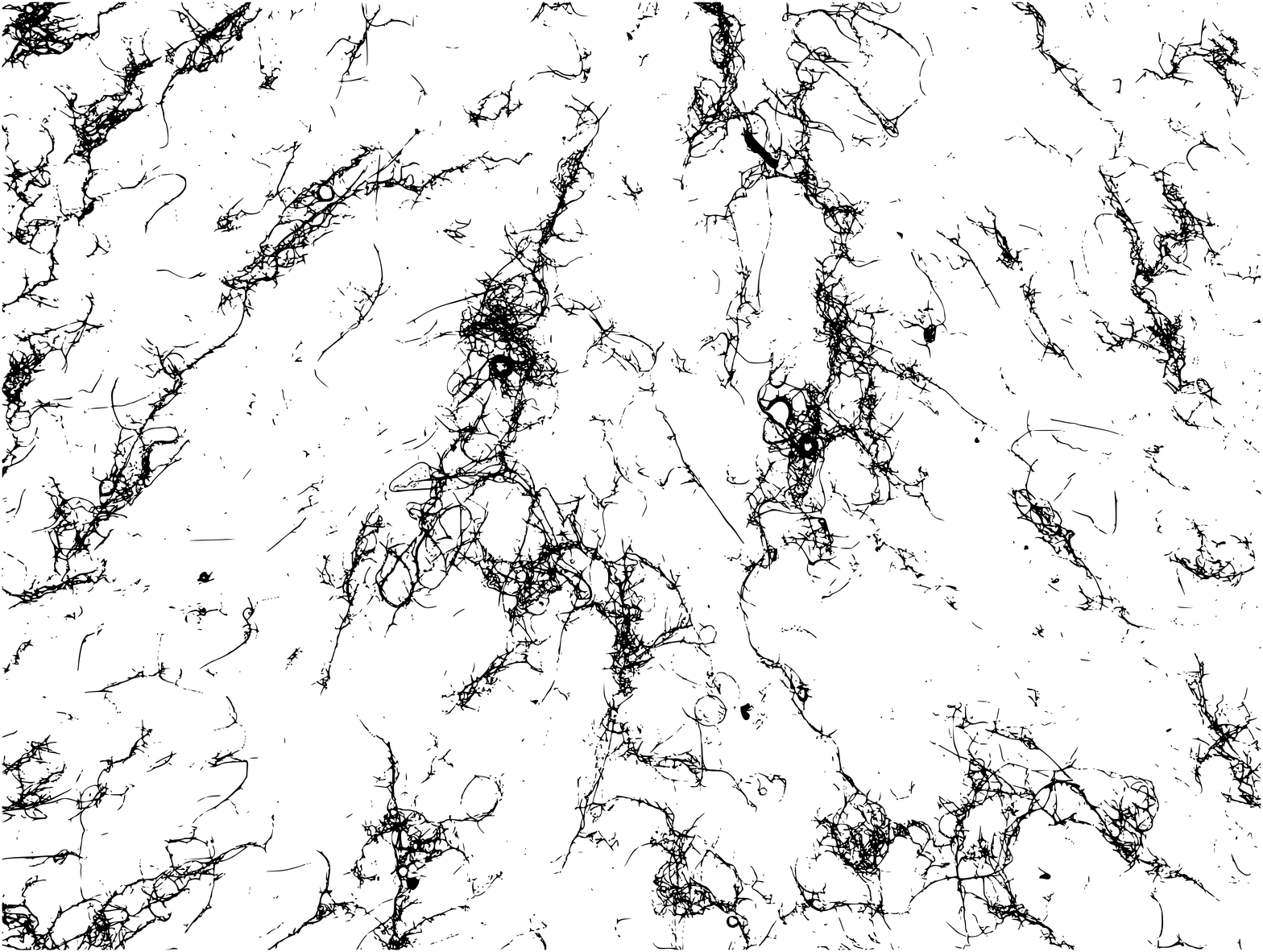}
\par\end{centering}
}\caption{\label{fig:Optimizing-the-desirable}Optimizing the desirable short
polymer fiber. (a) the truncated Gaussian prior on the butanol speed
extracted from the expert knowledge. (b) the comparison of maximal
percentage achieved at each iteration between different algorithms.
(c) the sample where we have achieved the maximal percentage fibers
whose length falls into the range of {[}50 150{]} microns.}
\end{figure*}

\subsection{Maximizing the Desirable SPF\label{subsec:Maximizing-the-Desirable fiber}}

\label{subsec:Maximizing-the-Desirable}Short polymer fibers (SPF)
are often used to coat natural fabrics to make them superior in many
aspects e.g. more resistive to pilling, improved water repellence
etc. Different types of fabrics generally require different sizes
of the fibers for optimal results. The fibers are produced by injecting
a polymer liquid through a high speed coagulant (e.g. butanol) flow
inside a specially designed apparatus \cite{sutti2014apparatus,LI_etal_ICDM2018}.
We aim to search for the sample with the maximal percentage fibers
with length falling into the range of {[}50 150{]} microns. There
are five parameters: geometric factors: channel width (mm), constriction
angle (degree), and device position (mm); and, flow factors: butanol
speed (cm/s), polymer concentration (ml/h) involved in the experimental
device, which construct 162 discrete combinations totally. Random
search is most straightforward. However, this experiment is very costly
and each takes at least half day, including component preparation,
experimental process and post analysis. BO becomes an ideal choice
for this expensive experimental design.

Several material experts have offered expert prior knowledge that
the optimal sample normally lies in the high value of butanol speed
in this case. In this experiment, butanol speed is discrete including
43, 68 and 95. Based on this weak prior knowledge, we have designed
a truncated Gaussian on the butanol speed (Figure \ref{subsec:Optimization-for-Synthetic}(a))
so that we can incorporate this knowledge to accelerate adaptive experimental
design through our developed framework. Since the search space is
discrete, we can directly compute the posterior for each combination
based on Eq.(\ref{eq:posterior2}) and then re-sample to suggest the
next evaluation point. We run algorithms for 5 times with different
3 initial samples. Figure \ref{subsec:Optimization-for-Synthetic}
(b) show the maximal value we achieved at each iteration. The results
indicate that the prior knowledge is effective on posterior sampling
for experimental design. We also demonstrate the optimal sample we
have obtained in \ref{subsec:Optimization-for-Synthetic} (c), which
has satisfied short polymer fiber experts.

\section{Conclusion\label{sec:conclusion}}

We are the first to present how to incorporate arbitrarily expert
prior knowledge about the global optimum location $\boldsymbol{x}^{*}$
to facilitate experimental design. We have used BO to perform sequential
experimental design. There are two difficulties to hinder this work
to be fully explored before: a) it is not clear how to represent the
prior knowledge about the optimum location in a tractable form although
the human can perceive, b) it is not clear how to incorporate it into
BO either in practice or theory. In this paper, we have addressed
both challenges. We have represented the prior knowledge about the
optimum location via a vague probability density function $\pi(\boldsymbol{x}^{*})$.
We infer the updated posterior conditioned on the prior $\pi(\boldsymbol{x}^{*})$.
Posterior sampling is then employed to suggest the next evaluation.
We demonstrate the efficiency of the proposed approach in several optimization tasks. Choosing a proper prior $\pi(\boldsymbol{x}^{*})$ is non-trivial
but our algorithm can also converge with a gross prior. It is interesting to develop algorithms to detect the misleading prior quickly in future. We expect
our work can provide insight towards incorporating some perceived
expert knowledge into experimental design or Bayesian optimization.

\section*{Descriptions}
\section*{Funding}
This  research  was  partially  funded  by the  Australian  Government  through  the  Australian  Research Council  (ARC).  Prof  Venkatesh  is  the  recipient  of  an  ARC Australian Laureate Fellowship (FL170100006). 
\section*{Conflicts of interest/Competing interests} 
The authors declare that they have no conflict of interest.
\section*{Availability of data and material} 
Most of data is public in this paper. 
\section*{Code availability}
https://tini.to/PJJH.
\section*{Authors' contributions} 
All authors contributed to manuscript writing. Problem formulation, algorithm development were performed by Cheng Li, Sunil Gupta, Santu Rana and Svetha Venkatesh. Experimental running and software coding were performed by Cheng Li. The experimental results were discussed by all authors.

\bibliographystyle{spmpsci}
\bibliography{MLjournal.bib}

\end{document}